\documentclass[acmtog,review=false,screen,acmauthoryear,timestamp]{acmart}

\usepackage{float}
\usepackage{amssymb}
\usepackage{enumitem}
\usepackage{soul}

\DeclareMathOperator*{\argmin}{argmin}

\DeclareMathOperator*{\nullspace}{Null}

\usepackage{booktabs} 

\usepackage[ruled]{algorithm2e} 

\SetAlFnt{\small}
\SetAlCapFnt{\small}
\SetAlCapNameFnt{\small}
\SetAlCapHSkip{0pt}
\IncMargin{-\parindent}

\acmJournal{TOG}
\acmVolume{9}
\acmNumber{4}
\acmArticle{39}
\acmYear{2018}
\acmMonth{3}

\setcopyright{none}

\acmDOI{0000001.0000001_2}

\setcitestyle{square}

\begin{document}
\title{MonoPerfCap: Human Performance Capture from Monocular Video}

%

\author{Weipeng Xu}
\email{wxu@mpi-inf.mpg.de}

\author{Avishek Chatterjee}
\author{Michael Zollh\"ofer}
\affiliation{%
	\institution{Max Planck Institute for Informatics}
}
\author{Helge Rhodin}
\affiliation{%
	\institution{EPFL}
}

\author{Dushyant Mehta}

\author{Hans-Peter Seidel}

\author{Christian Theobalt}
\affiliation{%
	\institution{Max Planck Institute for Informatics}
}




\begin{abstract}
%
We present the first marker-less approach for temporally coherent 3D performance capture of a human with general clothing from monocular video.
Our approach reconstructs articulated human skeleton motion as well as medium-scale non-rigid surface deformations in general scenes.
%
%
Human performance capture is a challenging problem due to the large range of articulation, potentially fast motion, and considerable non-rigid deformations, even from multi-view data.
Reconstruction from monocular video alone is drastically more challenging, since strong occlusions and the inherent depth ambiguity lead to a highly ill-posed reconstruction problem.
%
%
We tackle these challenges by a novel approach that employs sparse 2D and 3D human pose detections from a convolutional neural network using a batch-based pose estimation strategy.
Joint recovery of per-batch motion allows to resolve the ambiguities of the monocular reconstruction problem based on a low dimensional trajectory subspace.
In addition, we propose refinement of the surface geometry based on fully automatically extracted silhouettes to enable medium-scale non-rigid alignment.
%
%
We demonstrate state-of-the-art performance capture results that enable exciting applications such as video editing and free viewpoint video, previously infeasible from monocular video.
Our qualitative and quantitative evaluation demonstrates that our approach significantly outperforms previous monocular methods in terms of accuracy, robustness and scene complexity that can be handled.
\end{abstract}

%
%


%
%



\thanks{This work is supported by ERC StartingGrant ``CapReal'' (335545).}

\begin{teaserfigure}
  \includegraphics[width=\textwidth]{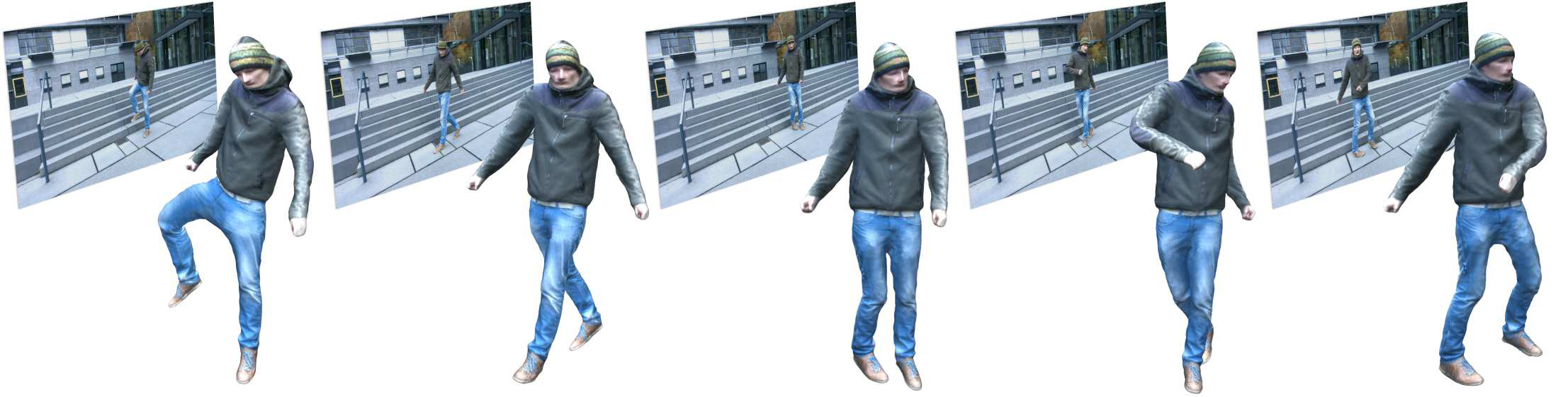}
  \caption{
  			We present the first marker-less approach for temporally coherent performance capture given just monocular video as input.
  			The reconstructed surface model captures the full articulated motion of the human body as well as medium-scale non-rigid deformations of the surface.
  	}
  \label{fig:intro}
\end{teaserfigure}

\maketitle



\section{Introduction}
%
%
Marker-free human performance capture has been a highly relevant and challenging research topic in the computer vision and computer graphics communities for the last decade.
Its goal is to track the motion of a moving subject, and reconstruct a temporally coherent representation of its dynamically deforming surface from unmodified videos.
%
%
Capturing the motion of humans is a challenging problem due to the high level of articulation, potentially fast motion and considerable non-rigid deformations.
%
%
A robust and highly accurate solution to this problem is a necessary precondition for a broad range of applications in not only computer animation, visual effects and free-viewpoint video, but also other fields such as medicine or biomechanics.
%
Especially, with the recent popularity of virtual reality (VR) systems and telepresence, there comes a rising demand of lightweight performance capture solutions.

In the literature of marker-less performance capture, multi-view methods~\cite{gall2009motion,vlasic2008articulated,de2008performance,liu2011markerless,cagniart2010free,bray2006posecut,brox2006high,brox2010combined,wu2013onset,mustafa2015iccv} have been well studied.
These techniques allow to obtain accurate results, but require expensive dense camera setups and controlled studios that are only available to a few technical experts.
With the recent commoditization of RGB-D sensors such as the Microsoft Kinect, many depth-based approaches~\cite{bogo2015detailed,shotton2011real} demonstrate the possibility of low cost performance capture with commodity hardware.
Even real-time tracking~\cite{zollhoefer2014deformable} and reconstruction~\cite{Newcombe_2015_CVPR,innmann2016volume} of general deforming objects has been demonstrated.
%
%
While reliable in controlled indoor settings, the active depth sensing modality of such devices hinders their application in direct sunlight.
Given their higher energy consumption, they are not as widely distributed as standard RGB cameras, especially on mobile devices.
Furthermore, depth-based approaches can not be applied to existing video footage, e.g.~from Youtube.

%
%
In this paper, we tackle the problem of human performance capture from monocular RGB video sequences, even outdoors with general background, to overcome the outlined limitations of depth cameras and multi-view setups.
Reconstruction from monocular video per se is a highly challenging and ill-posed problem due to strong occlusions and the lack of depth information.
Although several recent works target monocular tracking and reconstruction of specific, e.g. human faces~\cite{thies2016face,Garrido:2016}, and general~\cite{Garg_2013_CVPR,Russell2014,Yu_2015_ICCV} deformable surfaces, they only target objects undergoing relatively small deformations.
%
%
To the best of our knowledge, our approach is the first to handle the problem of automatic 3D full human body performance capture from monocular video input.
Similar to many existing performance capture approaches, ours employs an actor specific template mesh.
The deformation of the template mesh, obtained by image-based reconstruction prior to recording, is parameterized with a kinematic skeleton and a medium-scale deformation field.
%
%
Given this shape representation, we estimate the deformation of the actor for each frame in the input video, such that the deformed template closely matches the input frame.
The resulting algorithm allows us to generate a temporally coherent surface representation of the actor's full body performance.

%
%
In order to robustly capture the fast and highly articulated motion of the human body, we leverage 2D discriminative joint predictions from a convolutional neural network (CNN) as landmarks for registering the 3D skeleton to the image.
However, due to the lack of explicit depth input, 3D pose estimation suffers from a ``forward/backward flipping'' ambiguity at the revolute joints~\cite{sminchisescu2003kinematic}.
Therefore, the estimated 3D pose is often incorrect, even though the 2D projections of the skeleton joints accurately match the predictions.  
We tackle the flipping ambiguity with the help of a second CNN, which is trained to regress 3D joint positions from monocular images.
%
%
To further resolve the inherent depth ambiguity of the monocular reconstruction problem, we constrain the 3D poses in temporal space with a low dimensional linear trajectory subspace, which has proven effective in the context of non-rigid structure from motion~\cite{Hyun2015}.
%
In addition, we compute a non-rigid deformation field based on automatically extracted silhouettes to capture non-rigid surface deformation due to loose clothing, and accurately overlay the deformed template mesh onto the input image frames.

In summary, our monocular performance capture approach has the following main contributions:
%
%
\begin{itemize}
\item The first human 3D performance capture approach that relies only on monocular video input,
\item a combination of discriminative 2D and 3D detections and batch-based motion optimization to solve the inherent flipping ambiguities of monocular 3D pose estimation,
\item plausible recovery of non-rigid surface deformations with automatically extracted monocular silhouettes,
\item a benchmark dataset consisting of around 40k frames, which covers a variety of different scenarios.
\end{itemize}


\begin{figure*}[ht]
	\begin{center}
		\includegraphics[width=1\linewidth]{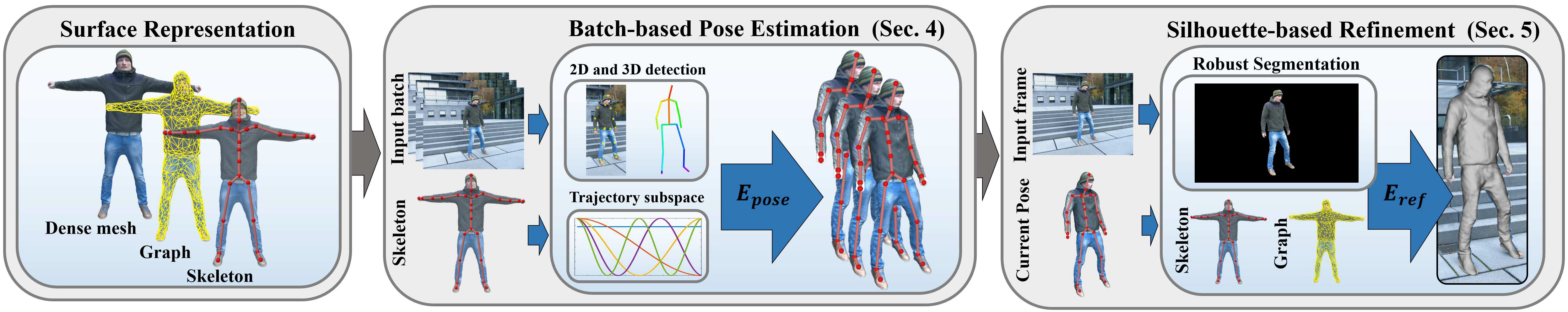}
	\end{center}
	\vspace{-0.1cm}
	\caption{
Given a monocular video and a personalized actor rig, our approach reconstructs the actor motion as well as medium-scale surface deformations.
The monocular reconstruction problem is solved by joint recovery of temporally coherent per-batch motion based on a low dimensional trajectory subspace.
Non-rigid alignment based on automatically extracted silhouettes is used to better match the input.
%
}
	\label{fig:pipeline}
\end{figure*}

\section{Related Work}
Performance capture has received considerable attention in computer vision and computer graphics.
Here, we focus on the works that are most related to our approach.

\paragraph{Multi-view Performance Capture}
Detailed surface geometry can be reconstructed using shape-from-silhouette and stereo constraints from multi-view footage \cite{matusik2000image,starck2007surface,waschbusch2005scalable}, and based on photometric stereo in a light stage \cite{vlasic2009dynamic}.
These model-free approaches require succeeding surface tracking to obtain temporal correspondence, e.g.~using \cite{cagniart2010free}.
Reconstructions with temporally consistent topology are obtained with model-based solutions, that deform an actor specific shape template to match silhouette and multi-view constraints \cite{Carranza:2003,de2008performance,bradley2008markerless}.
Incorporation of a kinematic skeleton model further regularizes the solution \cite{gall2009motion,vlasic2008articulated,liu2011markerless,wu2012full} and combined reconstruction and segmentation further improves accuracy \cite{bray2006posecut,brox2006high,brox2010combined,liu2011markerless,wu2013onset,mustafa2015iccv}.
The required actor model can be computed fully automatically using parametric models 
\cite{plankers2001tracking,balan2007detailed,sminchisescu2003estimating,hasler2010multilinear,anguelov2005scape,loper2014mosh,loper2015smpl,song20163d}, also in general environments \cite{rhodin2016general}.
These methods obtain high quality under controlled studio conditions, often with a green screen, but they do not work in general outdoor scenarios and the utilized multi-view and stereo constraints do not generalize to performance capture from a single consumer-level camera.

\paragraph{Depth-based Performance Capture}
Modern RGB-D sensors simultaneously capture synchronized color and depth at real-time frame rates.
This triggered the development of depth-based reconstruction approaches that fit articulated template models \cite{Li2009,Zhang2014,bogo2015detailed,Guo:2015,Helten:2013} that overcome many of the ambiguities of monocular RGB techniques.
%
%
Even real-time template-based non-rigid tracking \cite{zollhoefer2014deformable} and template-free reconstruction \cite{Newcombe_2015_CVPR,innmann2016volume,xu2015deformable} of general deforming scenes has been demonstrated.
Multi-view depth-based reconstruction obtains even higher accuracy and robustness~ \cite{dou2013scanning,Ye2012,collet2015high,dou2016fusion4d,wang2016capturing}.
While very reliable indoors, the active sensing modalities of consumer-level depth sensors hinders their application in direct sunlight, their high energy consumption is a drawback for mobile applications, and they are not yet as widely distributed as RGB cameras, which are already integrated in every smartphone.
Passive stereo depth estimation helps to overcome some of these limitations \cite{plankers2001tracking,wu2013onset}, but the required camera baseline is impractical for consumer-level applications and the quality of estimated depth is highly dependent on the amount of texture features in the reconstructed scene.

\paragraph{Sparse Skeletal Pose Reconstruction}
We make use of current advances in skeleton pose estimation, in particular from single views to bootstrap our surface reconstruction approach.
Motion capture solutions based on a generative image formation model require manual initialization \cite{wren1997pfinder} and pose correction.
\cite{wei2010videomocap} obtain high quality 3D pose from challenging sport video sequences using physical constraints, but require manual joint position annotations for each keyframe (every 30 frames).
Also simpler temporal priors have been applied \cite{sidenbladh2000stochastic,urtasun2005monocular,urtasun2006temporal}.
With recent advances in convolutional neural networks (CNNs), fully-automatic, high accuracy 2D pose estimation 
\cite{jain2014modeep,toshev2014deeppose,pishchulin2016deepcut,wei2016convolutional,newell2016stacked} is feasible from a single image. 
Lifting the 2D detections to the corresponding 3D pose is common \cite{taylor2000reconstruction,simo2012single,mori2006recovering,wang2014robust,akhter2015poseconditioned,li2015maximummargin,yasin2016dualsource}, but is a hard and underconstrained problem \cite{sminchisescu2003kinematic}.
\cite{bogo2016smpl} employ a pose prior based on a mixture of Gaussians in combination with penetration constraints.
The approach of \cite{zhou2015convex} reconstructs 3D pose as a sparse linear combination of a set of example poses.
Direct regression from a single image to the 3D pose is an alternative \cite{li2014threed,zhou2016deep,ionescu2014iterated,tekin2016structured,mehta2016monocular,pavlakos2016coarse}, but leads to temporally incoherent reconstructions.
 
%
Promising are hybrid approaches that combine discriminative 2D- \cite{elhayek2015efficient} and 3D-pose estimation techniques \cite{sminchisescu2006learning,rosales2006combining} with generative image formation models, but these approches require multiple views of the scene.
Recently, a real-time 3D human pose estimation approach has been proposed \cite{VNect_SIGGRAPH2017}, which also relies on monocular video input.
It is a very fast method, but does not achieve the temporal stability and robustness to difficult poses of our approach.
%
%
In contrast to this previous work, our method not only estimates the 3D skeleton more robustly, by leveraging the complimentary strength of 2D and 3D discriminative models, and trajectory subspace constraints, but also recovers medium-scale non-rigid surface deformations that can not be modeled using only skeleton subspace deformation.
We extensively compare to the approach of \cite{VNect_SIGGRAPH2017} in Sec.~\ref{sec:results}.
%

\paragraph{Dense Monocular Shape Reconstruction}
Reconstructing strongly deforming non-rigid objects and humans in general apparel given just monocular input is an ill-posed problem.
By constraining the solution to a low-dimensional space, coarse human shape can be reconstructed based on a foreground segmentation \cite{grest2005human,guan2009estimating,jain2010movie,chen2010inferring,zhou2010parametric,rogge2014garment}.
Still, these approaches rely on manual initialization and correction steps.
Fully automatic approaches combine generative body models with discriminative pose and shape estimation, e.g.~conditioned on silhouette cues \cite{sigal2007combined} and 2D pose \cite{bogo2016smpl}, but can also only capture skin-tight clothing without surface details.
The recent work of~\cite{MuVS:3DV:2017}, which fits a parametric human body model to the 2D pose detection and the silhouettes over time, has demonstrated compelling results on both multi-view and monocular data.
But again, their method is not able to model loose clothing. 
Model-free reconstructions are based on rigidity and temporal smoothness assumptions \cite{Garg_2013_CVPR,Russell2014} and only apply to medium-scale deformations and simple motions.
Template-based approaches enable fast sequential tracking \cite{Salzmann2011,Bartoli2015,Yu_2015_ICCV}, but are unable to capture the fast and highly articulated motion of the human body.
%
Automatic monocular performance capture of more general human motion is still an unsolved problem, especially if non-rigid surface deformations are taken into account.
Our approach tackles this challenging problem.


\section{Method Overview}

\begin{figure}[t]
	\begin{center}
		\includegraphics[width=0.7\linewidth]{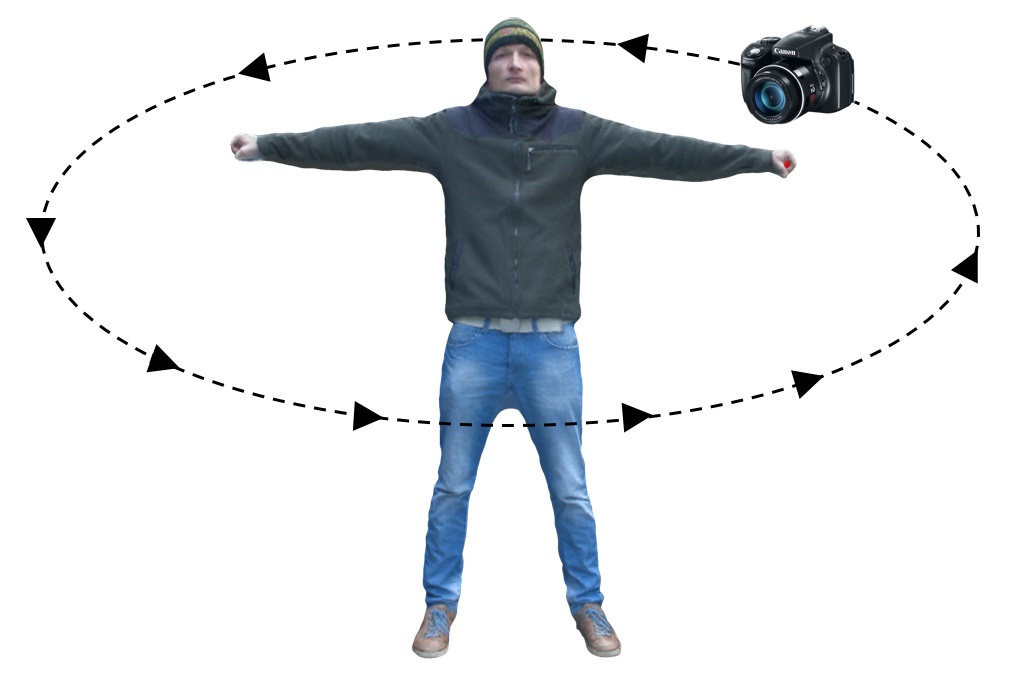}
	\end{center}
	\vspace{-0.3cm}
	\caption{
		Acquisition of a textured template mesh from handheld video footage of the actor in a static pose.
	}
	\vspace{-0.5cm}
	\label{fig:multiview}
\end{figure}

Non-rigid 3D reconstruction from monocular RGB video is a challenging and ill-posed problem, since the subjects are partially visible at each time instance and depth cues are implicit.
To tackle the problem of partial visibility, similar to many previous works, we employ 
a template mesh, pre-acquired by image based monocular reconstruction of the actor in a static pose.
When it comes to the scenario of capturing the full motion of a human body, the problem is even more challenging, due to the high degree of non-rigidity of humans, ranging from complex articulated motion to non-rigid deformations of skin and apparel.
We propose the first marker-less performance capture approach for temporally coherent reconstruction of 3D articulated human motion as well as 3D medium-scale surface deformations from just monocular videos recorded outside controlled studios.
To this end, we parameterize human motion based on a two level deformation hierarchy.
On the coarser level, the articulated motion is captured in skeleton deformation space.
On the finer level, a deformation field parameterized by an embedded deformation graph, models medium-scale non-rigid deformations of the surface.
%
%
Correspondingly, motion capture is performed in a coarse-to-fine manner, based on two subsequent steps, namely batch-based pose estimation (see Sec.~\ref{sec:batch}) and silhouette-based refinement (see Sec.~\ref{sec:refine}).
As shown in Fig.~\ref{fig:pipeline}, we first estimate the skeleton deformations in the input video, using a novel batch-based 3D pose estimation approach that exploits the discriminative 2D and 3D detections from trained CNNs and a linear trajectory subspace to obtain robust reconstruction.
%
The resulting temporally coherent reconstructions well reproduce articulated motion, but lack non-rigid surface deformations of the apparel and skin.
%
Consequently, there exists a noticeable misalignment between the skeleton-deformed model boundary and the image silhouette.
To alleviate this problem, we propose a surface refinement approach to better align the deformed model with automatically estimated actor silhouettes that are found by a model-guided foreground segmentation strategy (see Sec.~\ref{sec:silhouette_extraction}).
%

To obtain the person-specific template mesh, we first record a high-resolution video with a handheld camera orbiting around the actor standing in a T-pose, and uniformly sample $60$ images.
Afterwards, a triangulated surface with the corresponding texture is automatically reconstructed using the image based reconstruction software Agisoft Photoscan\footnote{http://www.agisoft.com}.
%
%

\section{Batch-based 3D Human Pose Estimation} \label{sec:batch}


We parameterize articulated human motion based on a low dimensional skeleton subspace \cite{Lewis:2000}.
The skeleton $\mathcal{S}=\{ \mathbf{t}, \mathbf{R}, \boldsymbol \Theta \}$ with $N_d=16$ joints $J_i$ is parameterized by the position $\mathbf{t} \in \mathbb{R}^3$ and rotation $\mathbf{R} \in  \mathbf{SO}(3)$ of its root joint, and $27$ angles stacked in $\boldsymbol \Theta \in \mathbb{R}^{27}$.
This leads to a $33$ dimensional deformation subspace.
The high-resolution actor mesh is rigged to the skeleton based on dual quaternion skinning \cite{Kavan:2007}.
%

	The skinning weights of our templates are automatically computed using Blender. For the skirt and long coat templates, we manually correct the skinning weights to reduce artifacts.

Given the monocular input video  $\mathcal{V} = \{I_f\}_{f=1}^{N}$ with $N$ image frames $I_f$, the goal of 3D pose estimation is to recover the skeleton parameters $\mathcal{S}_f$ for all input frames.
Since the problem of 3D human pose estimation is highly underconstrained given only a single RGB input frame $I_f$, we propose a novel batch-based approach that jointly recovers the motion for a continuous window in time:
\begin{equation}
\mathcal{B} = \{ \mathcal{S}_f ~|~ f_{start} \leq f \leq f_{end}\} ~,
\end{equation}
where $f_{start}$ specifies the index of the first and $f_{end}$ of the last frame included in the current batch.
In all our experiments, a constant batch size $|\mathcal{B}| = 50$ is used, and the input video is partitioned into a series of overlapping batches ($10$ frames overlap).
Each batch is processed independently and afterwards the per-batch skeleton reconstruction results are combined in the overlap region based on a linear blending function.

We phrase the problem of estimating the articulated motion of each batch $\mathcal{B}$ as a constrained optimization problem:
\begin{equation} \label{eq:opt}
\begin{aligned}
\mathcal{B}^* =~& \underset{\mathcal{B}}{\argmin}
& & E_{\textrm{pose}}(\mathcal{B})~, \\
& \text{subject to}
& &  \boldsymbol \Theta_{min} \leqslant \boldsymbol \Theta_{f} \leqslant \boldsymbol \Theta_{max}~, \\
&&&  \forall f \in [f_{start}, f_{end}]~,
\end{aligned}
\end{equation}
where the hard constraints on the per-frame joint angles $\boldsymbol\Theta_f$ are physically motivated and ensure the reconstruction of plausible human body poses by forcing the joint angles to stay inside their anatomical lower $\boldsymbol\Theta_{min}$ and upper $\boldsymbol\Theta_{max}$ bounds \cite{StollHGST11}.
The proposed batch-based pose estimation objective function $E_{\textrm{pose}}$ consists of several data fitting and regularization terms:
\begin{equation}
E_{\textrm{pose}}(\mathcal{B}) = \underbrace{E_{\textrm{2d}}(\mathcal{B}) + w_{3d}E_{\textrm{3d}}(\mathcal{B})}_{\textrm{data fitting}} + \underbrace{w_{d}E_{\textrm{d}}(\mathcal{B})}_{\textrm{regularization}}~.
\end{equation}
The data fitting terms ensure that the reconstructed motion closely matches the input:
A 2D joint alignment term $E_{\textrm{2d}}$ based on joint detections in image space and a 3D joint alignment term $E_{\textrm{3d}}$ based on regressed 3D joint positions.
The discriminative detections are obtained using CNNs that have been trained for 2D and 3D joint localization.
%
%
The motion of the skeleton is regularized on batch level by $E_{\textrm{d}}$ using a low dimensional trajectory subspace based on the discrete cosine transform.
This enforces the intra-batch motion to be temporally smooth, adds robustness against failed detections and further resolves depth ambiguity.
The weights $w_{\bullet}$ balance the relative importance of the different terms.
We provide more details in the remaining part of this section.

\paragraph{Discriminative Joint Alignment Terms}

For each input image $I_f$ and each of the $N_d=16$ joints $J_i$, we estimate the 2D joint position $\mathbf{d}_{f,i}^{\textrm{2d}}$ in image space and the 3D joint position $\mathbf{d}_{f,i}^{\textrm{3d}}$.
To this end, we use the Resnet~\cite{he2016deep} based CNN joint position regression method of~\cite{mehta2016monocular}, to which we add detections for the toes.
This better constrains the rotation of the feet and leads to higher quality reconstruction results.
%
%
Our 2D pose network is trained on the MPII Human Pose~\cite{andriluka2014human} and LSP~\cite{johnson2011learning} datasets, and the 3D pose network is fine-tuned from the 2D pose network on the H3.6M~\cite{ionescu2014iterated} and 3DHP~\cite{mehta2016monocular} datasets.
Our approach lifts the loose CNN detections to produce a coherent skeleton (parameterized by angles) and enforces constant bone length.
In contrast to previous works, e.g.~the 2D-to-3D lifting approach of \cite{zhou2015convex}, we incorporate both 2D and 3D constraints into our generative framework to allow for more robust pose estimation.
Our 2D joint alignment term is a re-projection constraint enforcing that the projected joint positions $J_i(\mathcal{S}_f)$ closely match the corresponding 2D detections $\mathbf{d}_{f,i}^{\textrm{2d}}$:
\begin{equation}
E_{\textrm{2d}}(\mathcal{B}) = \frac{1}{|\mathcal{B}|} \sum_{\mathcal{S}_f \in \mathcal{B}} { \frac{1}{N_d} \sum_{i=1}^{N_d}{\Big|\Big| \Pi\big( J_i( \mathcal{S}_f ) \big) - \mathbf{d}_{f,i}^{\textrm{2d}} \Big|\Big|_2^2}}~,
\end{equation}
where the mapping $\Pi : \mathbb{R}^3 \rightarrow \mathbb{R}^2$ implements the full perspective camera projection. 
We apply this constraint independently for every frame of $\mathcal{B}$.
In addition, we propose a 3D joint alignment term based on the regressed 3D joint positions $\mathbf{d}_{f,i}^{3d}$:
\begin{equation}
E_{\textrm{3d}}(\mathcal{B}) = \frac{1}{|\mathcal{B}|} \sum_{\mathcal{S}_f \in \mathcal{B}} { \frac{w_f}{N_d} \sum_{i=1}^{N_d}{\Big|\Big| J_i( \mathcal{S}_f ) - \big(\mathbf{d}_{f,i}^{\textrm{3d}} + \mathbf{t}_f\big) \Big|\Big|_2^2}}~.
\end{equation}
Since the 3D joint detections $\mathbf{d}_{f,i}^{3d}$ are normalized for a skeleton with average bone length and are predicted relative to the root joint, rather than in camera space, they have to be rescaled to match the actor model, and mapped to their corresponding camera space position based on an unknown per-frame global translation $\mathbf{t}_f$.
In order to prune frames with low 3D detection confidence, we measure the per-frame PCK error~\cite{toshev2014deeppose} $\textrm{PCK}_f$ between the 2D joint detections and the projected 3D detections and apply a per-frame binary weight $w_f$ to the 3D data term:
\begin{equation}
w_f =
\begin{cases}
1 & \textrm{ if } \textrm{PCK}_f < thres_{pck}, \\
0 & \textrm{ else.}
\end{cases}
\end{equation}
where $thres_{pck}=0.4$ is an empirically determined, but constant, distance threshold.
Note, the 2D detections are always included in the optimization, since they have a higher reliability.

%

\paragraph{Batch-based Motion Regularization}

Up to now, all poses $\mathcal{S}_f$ are temporally independent and sometimes inaccurate, since the monocular reconstruction problem is highly underconstrained.
To alleviate this problem, we impose temporal smoothness by forcing the trajectory of each skeleton parameter to lie on a low dimensional linear subspace.
Specifically, we couple all pose estimates $\mathcal{S}_f \in \mathcal{B}$ by minimizing the distance to a $K=8$ dimensional linear subspace $\textbf{DCT} \in \mathbb{R}^{ K \times |\mathcal{B}|}$ \cite{Hyun2015} spanned by the $K$ lowest frequency basis vectors of the discrete cosine transform (DCT):
\begin{equation} \label{eq:dct}
E_{\textrm{d}}(\mathcal{B}) = \frac{1}{|\mathcal{B}|} \Big|\Big| \boldsymbol{\Lambda}  \textbf{S}_{\mathcal{B}}  \nullspace(\textbf{DCT})  \Big|\Big|_{F}^{2}~.
\end{equation}
Here, $\nullspace(\textbf{DCT})$ denotes the nullspace of the $\textbf{DCT}$ matrix, and the matrix $\mathbf{S}_\mathcal{B}$ stacks all parameters $\mathcal{S}_f$ of the current batch:
\begin{equation}
\mathbf{S}_\mathcal{B} = \begin{bmatrix}  \mathcal{S}_{f_{start}}~, & \dots &,~ \mathcal{S}_{f_{end}} \\ \end{bmatrix} \in \mathbb{R}^{  |\mathcal{S}|\times|\mathcal{B}|}~.
\end{equation}
%
%
The diagonal matrix $\boldsymbol{\Lambda} = \textrm{diag}([\lambda_{\mathbf{t}}, \lambda_{\mathbf{R}}, \lambda_{\boldsymbol\Theta}])$ balances the motion smoothness of the global translation, rotation and joint angle components.
In all our experiments, $\lambda_{\mathbf{t}}= 1 \cdot \mathbf{1}_3$, $\lambda_{\mathbf{R}} = 600 \cdot \mathbf{1}_3$ and $\lambda_{\boldsymbol\Theta} = 600 \cdot \mathbf{1}_{27}$, where $\mathbf{1}_k$ is the $k$-dimensional row vector of ones. $\|\cdot\|_F$ denotes the Frobenius norm.
%

\paragraph{Initialization and Optimization} 
The optimization problem proposed in Eq.~\ref{eq:opt} is non-linear due to the involved camera projection and the hierarchical parameterization of articulated motion based on joint angles.
%
%
We solve this constrained non-linear least squares optimization problem using the Levenberg Marquardt (LM) algorithm provided by Ceres\footnote{http://ceres-solver.org}.
%

Since the optimization problem is non-convex, LM requires an initialization close to the global optimum for convergence.
To this end, we resort to a per-frame initialization strategy by finding the $\mathcal{S}_f$ that minimize a joint alignment energy function $E_{\textrm{2d}}+E_{\textrm{3d}}$.

\section{Silhouette-based Refinement} \label{sec:refine}

As mentioned before, our batch-based pose optimization does not capture non-rigid surface deformation due to apparel and skin, and thus leads to misalignments between the skeleton-deformed template mesh and the input images, particularly at the boundaries.
%
%
%
To alleviate this problem, we propose a pose and surface refinement method based on automatically extracted silhouettes.

\begin{figure}[t]
	\begin{center}
		\includegraphics[width=1\linewidth]{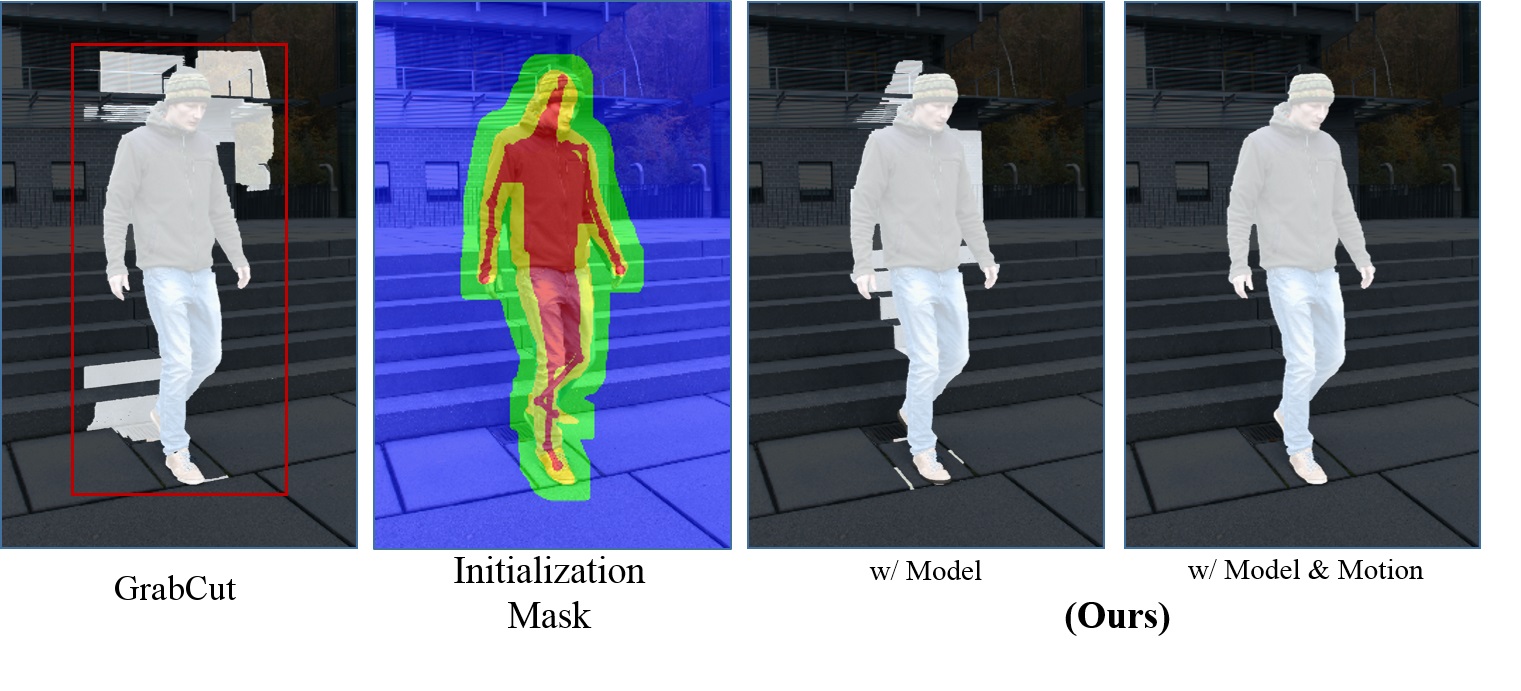}
	\end{center}
	\vspace{-0.6cm}
	\caption{
Our fully automatic model-based initialization (middle/left) significantly improves the segmentation (middle/right) compared to manual initalization based on a bounding box (left).
In addition, we use motion cues to further improve the results (right).
	}
	\label{fig:segmentation}
\end{figure}

\subsection{Automatic Silhouette Extraction} \label{sec:silhouette_extraction}

Given an input frame $\mathbf{I}_f$, we estimate the silhouette of the actor through a foreground segmentation method based on GrabCut \penalty700\ \cite{rother2004grabcut}.
GrabCut requires a user-specified initialization $\mathbf{T} = \{\mathbf{T}_b, \mathbf{T}_{ub}, \mathbf{T}_{uf}, \mathbf{T}_f \}$, where $\mathbf{T}_f$ and $\mathbf{T}_b$ denote the known foreground and background masks and the segmentation is computed over the remaining uncertain foreground $\mathbf{T}_{uf}$ and background $\mathbf{T}_{ub}$ regions.
The original GrabCut interactively initializes the masks based on a user-specified bounding box $\mathbf{B}$ and sets $\mathbf{T}_{uf} = \mathbf{T}_{ub} = \mathbf{B}$, $\mathbf{T}_b = \overline{\mathbf{B}}$, and $\mathbf{T}_f = \emptyset$.
In contrast, we propose a fully automatic initialization strategy for $\mathbf{T}$ based on the skeleton parameters $\mathcal{S}_f$ obtained by our batch-based pose estimation.
%
%
%
%
%
To this end, we first rasterize the skeleton and the deformed dense actor template $V(\mathcal{S}_f)$ to obtain two masks $\mathbf{R}$ and $\mathbf{M}$ respectively. 
Then we set the masks $\mathbf{T}$ as follows:
\begin{equation}
	\begin{aligned}
	&\mathbf{T}_{f} = R \cup erosion(\mathbf{M}), \\
	&\mathbf{T}_b = \overline{dilation(\mathbf{M})},\\
	&\mathbf{T}_{uf} = \mathbf{M} - \mathbf{T}_{f}, \\
	&\mathbf{T}_{ub} = dilation(\mathbf{M}) - M,
	\end{aligned}
\end{equation}
where $erosion(\cdot)$ and $dilation(\cdot)$ denote the image erosion and dilation operator.
In Fig.~\ref{fig:segmentation} (Initialization Mask), the masks  $\{\mathbf{T}_b, \mathbf{T}_{ub}, \allowbreak \mathbf{T}_{uf}, \mathbf{T}_f \}$ are illustrated in red, blue, yellow and green.
We show our robust model-based segmentation result on the complete sequence in the accompanying video.

To improve the robustness of our segmentation method, we extend the original GrabCut objective function, by incorporating motion cues.
Specifically, we extract the temporal per-pixel color gradients between adjacent frames and encourage neighboring pixels with small temporal gradients to belong to the same region.
As shown in  Fig.~\ref{fig:segmentation}, our model-based strategy and the extension with motion cues lead to fully automatic and significantly improved segmentation.

\subsection{Silhouette-based Pose Refinement}\label{Silhouette-based_Pose_Refinement}
Our silhouette-based pose refinement is performed in an Iterative Closest Point (ICP) manner.
In an ICP iteration, for each boundary point of the projected surface model, we search for its closest point on the image silhouette that shares a similar normal direction.
Then we refine the pose by solving the following non-linear least squares optimization problem:
\begin{equation}
E_{\textrm{ref}}(\mathcal{S}_f) = \underbrace{E_{\textrm{con}}(\mathcal{S}_f)}_{\textrm{data fitting}} + \underbrace{w_{\textrm{stab}} E_{\textrm{stab}}(\mathcal{S}_f)}_{\textrm{regularization}}~,
\end{equation}
where $E_\textrm{con}$ aligns the mesh boundary with the input silhouette, $E_{\textrm{stab}}$ constrains the solution to stay close to the batch-based results and $w_{\textrm{stab}}$ balances the importance of the two terms.
We initialize the iterative pose refinement with the batch-based pose estimates, and typically perform 3 iterations.
 
\paragraph{Silhouette Alignment Constraint}

The closeness of corresponding points is enforced as follows:
\begin{equation}
E_{\textrm{con}}(\mathcal{S}_f) = \frac{1}{|\mathbf{S}|} \sum_{k \in \mathbf{S}}{\Big[ \mathbf{n}_k^T \cdot \Big( \Pi\big( \mathbf{v}_k(\mathcal{S}_f) \big) - \mathbf{s}_k \Big) \Big]^{2}}~,
\end{equation}
where $\mathbf{S}$ is the boundary of the actor model, $\mathbf{v}_k$ the position of vertex $k$ and $\mathbf{s}_k \in \mathbb{R}^2$ the corresponding silhouette point in the image with 2D normal $\mathbf{n}_k$.

\paragraph{Pose Stabilization Constraint}
We enforce the refined skeleton pose to be close to its initialization based on the following soft-constraint:
\begin{equation}
E_{\textrm{stab}}(\mathcal{S}_f) = \frac{1}{N_d} \sum_{i=1}^{N_d}{ \Big|\Big| J_i(\mathcal{S}_f) - J_i(\hat{\mathcal{S}}_f) \Big|\Big|_{2}^{2}}~,
\end{equation}
where $\hat{\mathcal{S}}_f$ are the joint angles after batch-based pose estimation and $J_i(\cdot)$ computes the 3D position of joint $J_i$.

\begin{figure}[t]
	\begin{center}
		\includegraphics[width=1.0\linewidth]{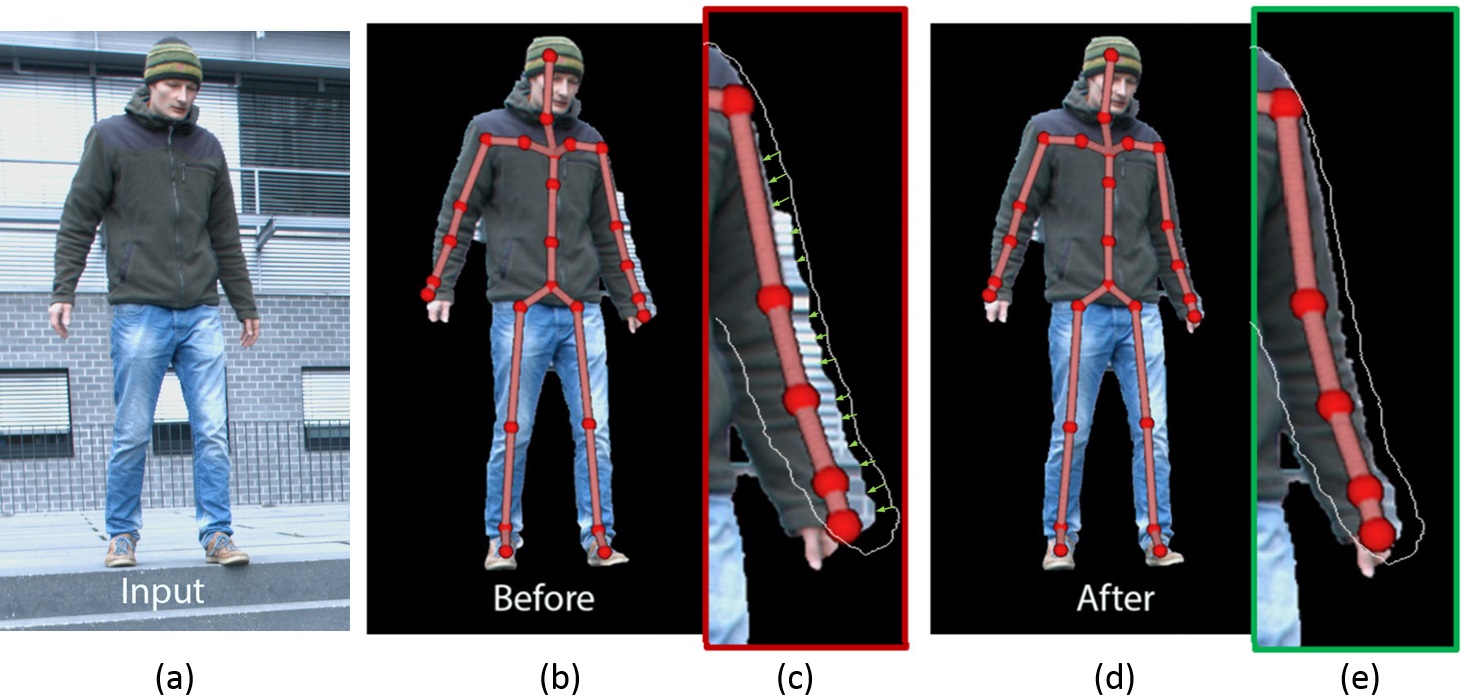}
	\end{center}
	\caption{
		Our silhouette-based pose refinement improves both pose estimation and the silhouette segmentation.
		The error in pose estimation cause inaccurate background subtraction near the left arm (b,c).
		Our silhouette pose refinement pulls the mesh (in white) to the silhouette (c), and therefore moves the arm skeleton leftwards to the correct position (d).
		The silhouette segmentation is significantly refined after the second silhouette extraction based on the refined pose (e).
	}
	\label{fig:pose_refine}
\end{figure}

After the iterative pose refinement, we perform the silhouette extraction of Sec.~\ref{sec:silhouette_extraction} for a second time, to further improve the segmentation.
As shown in Fig.~\ref{fig:pose_refine}, our iterative pose refinement not only improves the pose estimates, but also significantly increases the accuracy of the silhouette segmentation, which allows for the more accurate non-rigid surface alignment of Sec.~\ref{Silhouette-based_Surface_Refinement}.

\subsection{Silhouette-based Non-Rigid Surface Refinement}\label{Silhouette-based_Surface_Refinement}

Given the silhouette segmentation improved by iterative pose refinement, we perform a surface refinement step based on a medium-scale deformation field to closely align the model to the extracted image silhouettes.
This captures the non-rigid surface deformations of apparel and skin that are visible in the silhouette outline.
Refinement of the interior is hard due to a potential lack of strong photometric cues.

We parameterize the medium-scale warp field using an embedded deformation graph \cite{Sumner:2007}.
The deformation graph $\mathcal{D}$, consisting of $M\approx 1000$ nodes, is generated from the template mesh using a uniform mesh decimation/simplification strategy.
We assign a radius of influence for each deformation node by computing the maximum geodesic distance to its connected graph nodes. 
Each node defines a local warp field $W_i$ that rotates $\mathbf{R}_i \in \mathbf{SO}(3)$ and translates $\mathbf{t}_i \in \mathbb{R}^3$ points $\mathbf{x} \in \mathbb{R}^3$ in the surrounding space:
\begin{equation}
W_i(\mathbf{x}) = \mathbf{R}_i (\mathbf{x} - \hat{\mathbf{g}}_i)  +  \hat{\mathbf{g}}_i + \mathbf{t}_i~,
\end{equation}
where $\hat{\mathbf{g}}_i\in \mathbb{R}^3$ is the canonical position of node $i$, computed with the result of the pose refinement. 
We refer to the graph and its associated degrees of freedom as:
\begin{equation}
\mathcal{D} = \{ (\mathbf{R}_i, \mathbf{t}_i) | i \in [0, M)\}~.
\end{equation}
We apply the medium-scale deformation field to the dense actor model by linear blending of the per-node warp fields:
\begin{equation}
\mathbf{v}_i = W(\hat{\mathbf{v}}_i) = \sum_{k \in \mathbf{F}_i}{ b_{i, k}(\mathbf{x}) \cdot  W_k(\hat{\mathbf{v}}_i) }~.
\end{equation}
Here, $\mathbf{v}_i \in \mathbb{R}^3$ is the deformed vertex position, $\hat{\mathbf{v}}_i \in \mathbb{R}^3 $ is the canonical position of vertex $i$ and $\mathbf{F}_i$ is the set of deformation nodes that influence vertex $i$.
%
%
We compute the blending weights $b_{i,k}$ based on an exponential distance falloff and make them a partition of unity.

Given the embedded deformation graph, our silhouette-based surface refinement is expressed as the following optimization problem:
\begin{equation}
E_{\textrm{surf}}(\mathcal{D}) =  \underbrace{E_{\textrm{con}}(\mathcal{D})}_{\textrm{data fitting}} + \underbrace{w_{\textrm{arap}}E_{\textrm{arap}}(\mathcal{D})}_{\textrm{regularization}}~.
\end{equation}
Here, $E_{\textrm{con}}$ is the silhouette alignment term, $E_{\textrm{arap}}$ an as-rigid-as-possible regularization term \cite{Sorkine:2007} and $w_{\textrm{arap}}$ balances the two terms.


Our silhouette alignment term $E_{\textrm{con}}$ encourages the actor model to tightly align with the input silhouette:
\begin{equation}
E_{\textrm{con}}(\mathcal{D}) = \frac{1}{|\mathbf{S}|} \sum_{k \in \mathbf{S}}{\Big[ \mathbf{n}_k^T \cdot \Big( \Pi\big( \mathbf{v}_k(\mathcal{D}) \big) - \mathbf{s}_k \Big)\Big]^{2}}~,
\end{equation}
where $\mathbf{S}$ is the model silhouette, $\mathbf{v}_k$ the position of vertex $k$ and $\mathbf{s}_k \in \mathbb{R}^2$ its corresponding silhouette point with normal $\mathbf{n}_k \in \mathbb{R}^2$.

The as-rigid-as-possible term regularizes the non-rigid surface deformation of the graph nodes:
\begin{equation}
E_{\textrm{arap}}(\mathcal{D}) = \frac{1}{M} \sum_{i=1}^{M}{ \sum_{j \in \mathcal{N}_i} {\Big|\Big|  (\mathbf{g}_i - \mathbf{g}_j) - \mathbf{R}_i (\hat{\mathbf{g}}_i - \hat{\mathbf{g}}_j)  \Big|\Big|_{2}^{2}}}~.
\end{equation}
Here, $\mathbf{g}_i = W_i(\hat{\mathbf{g}}_i) = \hat{\mathbf{g}}_i + \mathbf{t}_i$ is the deformed position of node $\hat{\mathbf{g}}_i$ and $\mathcal{N}_i$ is its $1$-ring neighbourhood.

Similar to pose refinement (see Sec.~\ref{Silhouette-based_Pose_Refinement}), we perform surface refinement in an ICP-like manner.
To this end, we iterate the correspondence search and the model alignment step two times.
We initialize the optimization problem based on the pose refinement result and minimize $E_\textrm{surf}$ using the LM algorithm.
The final results are obtained by temporally smoothing the per-frame results based on a centered window of $5$ frames.


	\begin{figure}[t]
	\begin{center}
		\includegraphics[width=0.95\linewidth]{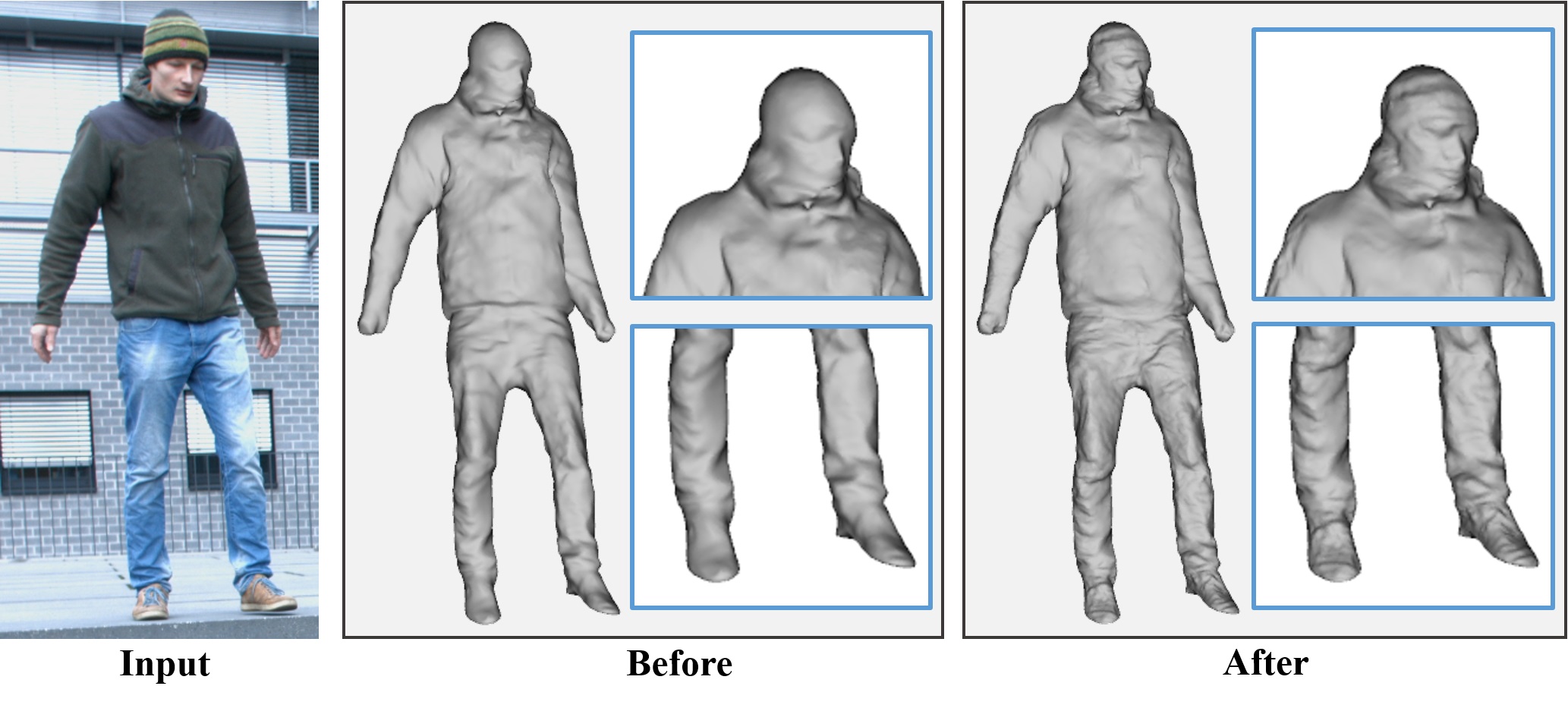}
	\end{center}
	\vspace{-0.3cm}
	\caption{
Our reconstruction results can optionally be refined to add fine-scale surface detail.
	}	
	\label{fig:refine}
\end{figure}

\begin{figure}[t]
	\centering
	\includegraphics[width=\linewidth]{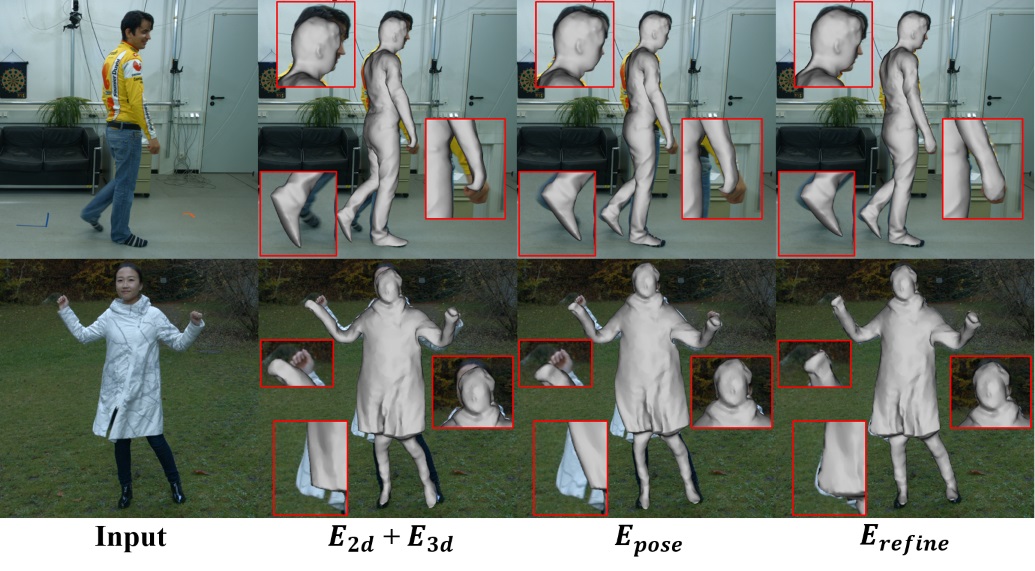}
	\vspace{-0.6cm}
	\caption{
Qualitative evaluation of components:
The batch-based pose optimization $E_{\textrm{pose}}$ significantly improves alignment over the discriminative energy $E_{\textrm{2d}} + E_{\textrm{3d}}$.
Note the improved rotation of the feet.
Residual non-rigid deformations are compensated via surface-based silhouette refinement $E_{\textrm{refine}}$.
}
	\label{fig:components_qualitative}
\end{figure}

\renewcommand{\dbltopfraction}{.9}
\renewcommand{\textfraction}{.1}

\begin{figure*}[ht]
	\begin{center}
		\includegraphics[width=1.0\linewidth]{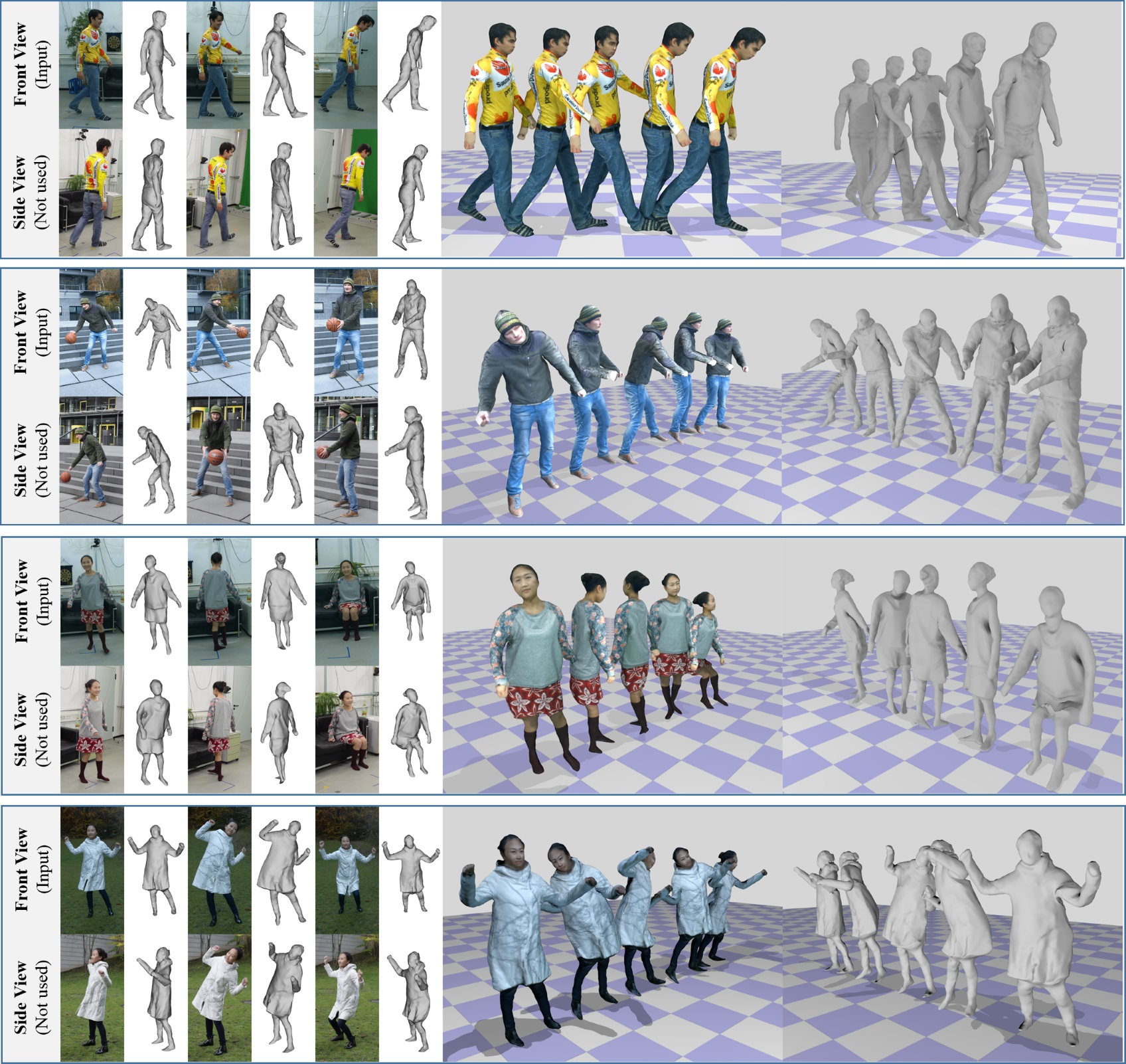}
	\end{center}
	\vspace{-0.2cm}
	\caption{
Qualitative Results:
We demonstrate compelling monocular performance capture results on a large variety of challenging scenes (left) that span indoor and outdoor settings, natural and man-made environments, male and female subjects, as well as tight and loose garments.
Our reconstructions match the real world even when viewed from the side.
Note, the side views are just used for reference, and are not used as input to our approach.
The reconstructions obtained by our approach are the basis for free-viewpoint video (right). For more results on our benchmark dataset we refer to the accompanying video.
}
	\vspace{-0.3cm}
	\label{fig:qualitative}
\end{figure*}

\section{Results} \label{sec:results}

%
%
In all experiments, we use the following empirically determined parameters to instantiate our energy functions:
$w_{3d} = 0.1$, $w_{p} = 0.1$, $w_{d} = 50$, $w_{stab} = 0.06$, and $w_{arap}$ is set to $0.6$ and $0.2$ for the two ICP iterations respectively.
Our approach proved robust to the specific choice of parameters, and thus we use this fixed set in all experiments.
%
We performed all experiments on a desktop computer with a $3.6$GHz Intel Xeon E5-1650 processor.
Our unoptimized CPU code requires approximately $1.2$ minutes to process one input frame.
This divides into 10 seconds for batch-based pose estimation and 1 minute for surface refinement.
We believe that the runtime of our approach can be greatly improved based on recent progress in data-parallel optimization~\cite{zollhoefer2014deformable}.
In the remaining part of this section, we first discribe the proposed benchmark dataset, then present our qualitative results, then evaluate all components of our approach, and finally compare to state-of-the-art monocular and multi-view approaches quantitatively.

\subsection{Benchmark dataset}

To evaluate our monocular performance capture approach for a variety of scenarios, we propose a benchmark dataset consisting of 13 sequences (around 40k frames in total), which divides into the following subsets:
1) We captured $8$ video sequences at $30$Hz, which cover a variety of different scenarios including indoor and outdoor settings, handheld and static cameras, natural and man-made environments, male and female subjects, as well as tight and loose garments.
2) To further increase the diversity of human motions of our benchmark dataset, we captured 40 additional actions, including daily actions such as walking, jumping as well as highly challenging ones such as rolling, kicking and falling. Each action is repeated multiple times by 3 subjects. In total, this leads to 120 video clips in 3 long video sequences, 7 minutes each.

3) In addition, we included two sequences from prior works, \cite{robertini2016model} and \cite{wu2013onset}, in our benchmark dataset.
These two sequences provide accurate surface reconstruction from multiview images, which can be used as ground truth for quantitative evaluation.

In order to evaluate our results in a different view from the one used as input, we also captured a side view using a second camera, and calibrated the extrinsic parameters for both cameras.
We also provide manually labeled silhouettes for a subset of the benchmark, which are used to compute silhouette overlap as a metric for quantitative evaluation of the performance capture results.
The benchmark dataset and our performance capture results will be made publicly available.

\subsection{Qualitative Results}

Our qualitative results are shown in Fig.~\ref{fig:qualitative}. We refer to the accompanying video for the complete results on our entire benchmark dataset.

\begin{figure}[t]
	\includegraphics[width=\linewidth]{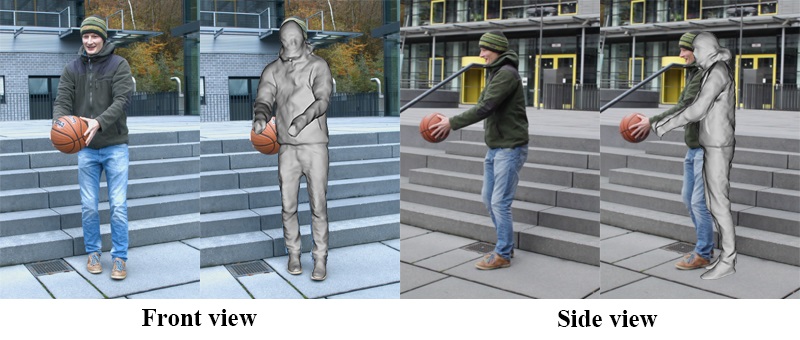}
	\vspace{-0.6cm}
	\caption{
Despite a small depth offset between the reconstruction and the actor in the reference view (not used for tracking) due to the remaining monocular depth ambiguity, our approach is able to accurately recover the deforming pose and shape.
}
	\label{fig:sideview}
\end{figure}

Our approach accurately captures the performance of the actors and obtains temporally coherent results on all test sequences, even for challenging motions with 360 degrees of rotation and sitting down on a sofa.
Even continuous interactions with objects, e.g., the basketball, is allowed.
To the best of our knowledge, no previous monocular approach could handle such scenarios.
%
%
%
As shown in Fig.~\ref{fig:sideview}, from the reference view, in spite of a small offset between the projected meshes and the actor due to some ambiguities in monocular depth estimation, our approach is able to accurately recover the full 3D deforming pose and the shape of the actor.
The textured spatio-temporal reconstructions are the basis for free-viewpoint videos and can be rendered from arbitrary viewpoints, see Fig.~\ref{fig:qualitative} (right).
Our reconstructed models also allow us to employ shading-based refinement using estimated lighting \cite{wu2013onset} to recover fine-scale surface detail, see Fig.~\ref{fig:refine}, but this is not the focus of our work.
We also tested our approach on the 3 long video sequences, 7 minutes each.
Our approach is able to continuously track each sequence in one go without restarting.
This demonstrates its robustness and generality.
Our complete results on all 3 sequences, which contain very challenging motions that bring our approach to its limits, are provided in the second supplementary video.
Furthermore, one of the sequences in our benchmark dataset is captured with a handheld camera, demonstrating the effectiveness of our approach even for non-static cameras.

\subsection{Evaluation of Algorithmic Components}

\begin{figure}[t]
		\includegraphics[width=\linewidth]{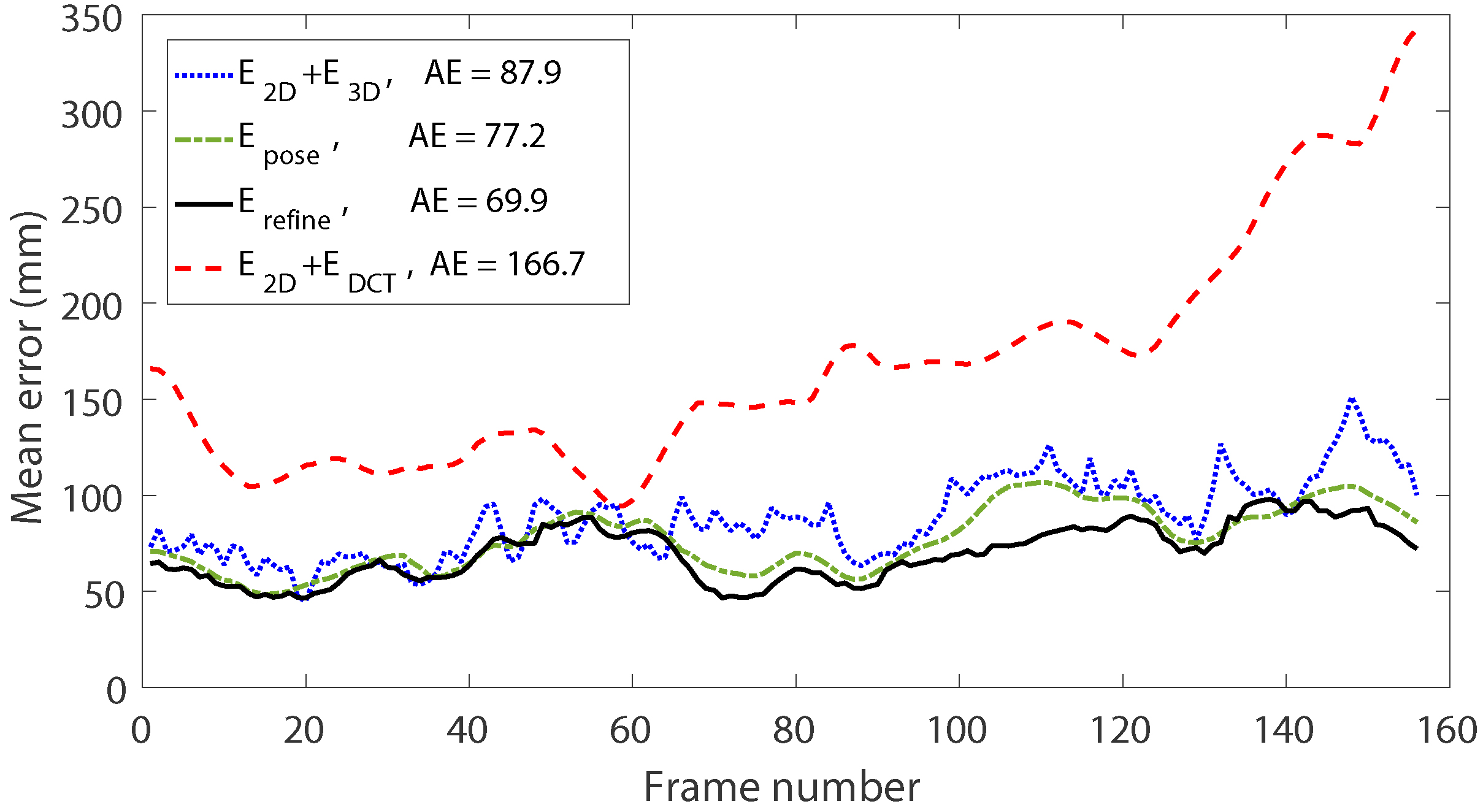}
		\vspace{-0.5cm}
		\caption{
Quantitative evaluation of components:
All steps of our approach improve the surface reconstruction error (in millimeters). The average error (AE) over all frames is given in the legend. }
%
		\label{fig:components_quantitative}
		\vspace{-0.5cm}
\end{figure}

The three main steps of our approach are:
Frame-to-frame 3D skeleton pose initialization based on the 2D/3D predictions ($E_{\textrm{2d}} + E_{\textrm{3d}}$), batch-based pose estimation ($E_{\textrm{pose}}$) and silhouette-based pose and surface refinement ($E_{\textrm{refine}}$).
We demonstrate the importance of all steps by comparing the results qualitatively in Fig.~\ref{fig:components_qualitative}.
While the joint detection based initialization ($E_{\textrm{2d}} + E_{\textrm{3d}}$) yields plausible results, the following batch-based pose estimation step ($E_{\textrm{pose}}$), which exploits temporal smoothness, improves the overlay and also removes the temporal jitter of the temporally incoherent 2D/3D CNN joint detections (see the accompanying video).
Note, in contrast to our coherent skeletal pose reconstruction, the CNN joint detections do not enforce a temporally constant bone length.
Furthermore, our silhouette-based refinement step, which produces our final results, significantly improves the overlay.
%
%
In addition, we also quantitatively evaluate the contribution of each component.
To this end, we made use of the multi-view performance capture method described in~\cite{robertini2016model}, which has demonstrated convincing results in capturing both pose and surface deformations in outdoor scenes.
We select a single view of one of their multi-view sequences (\emph{Pablo} sequence) as a test set, and use their results as ground truth for quantitative evaluation. 
%
As shown in Fig.~\ref{fig:components_quantitative} (average errors are given in the legend), our three main steps gradually improve the per-frame surface-to-surface mean error, while the complete approach ($E_{\textrm{refine}}$) has the lowest error over almost all frames.
%
%
%
%
%
Note that, for this evaluation, we aligned the reconstruction to the ground truth with a translation to eliminate the global depth offsets shown in Fig.~\ref{fig:sideview}.

\begin{figure}[t]
	\begin{center}
		\includegraphics[width=1\linewidth]{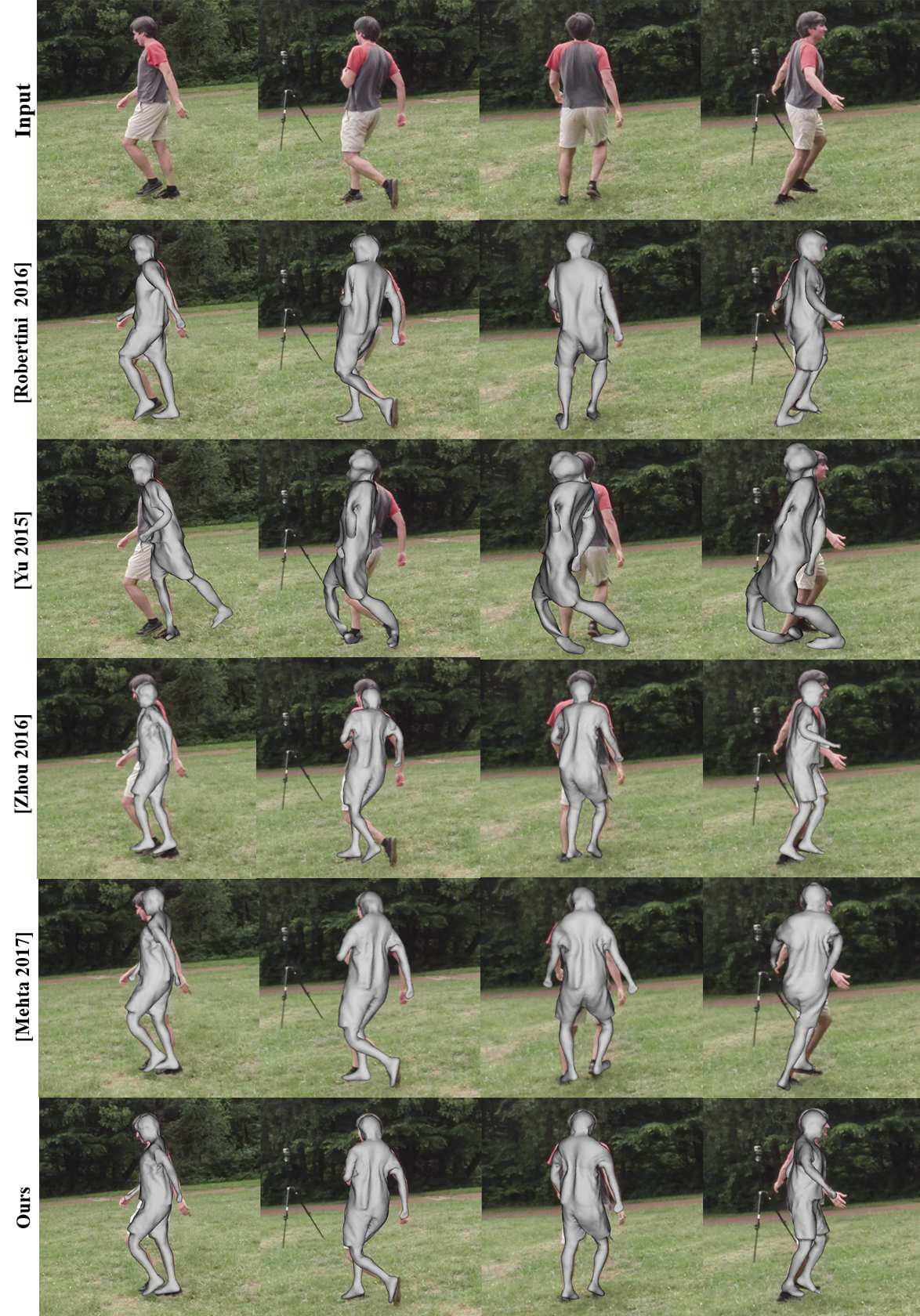}
	\end{center}
	\caption{
		Qualitative shape comparison on the \emph{Pablo} sequence.
		Our approach obtains comparable quality as the multi-view approach of \protect\cite{robertini2016model} (8 cameras) and drastically outperforms the template-based monocular tracking approach of \protect\cite{Yu_2015_ICCV}. In comparison to rigging the template model with respect to the 3D pose estimation results of \protect\cite{VNect_SIGGRAPH2017} and~\protect\cite{zhou2016sparseness}, our approach yield more accurate results with less artifacts that better overlay with the input.
	}
	\label{fig:pablo_yu}
	\vspace{-0.5cm}
\end{figure}

\begin{figure}[t]
	\begin{center}
		\includegraphics[width=1\linewidth]{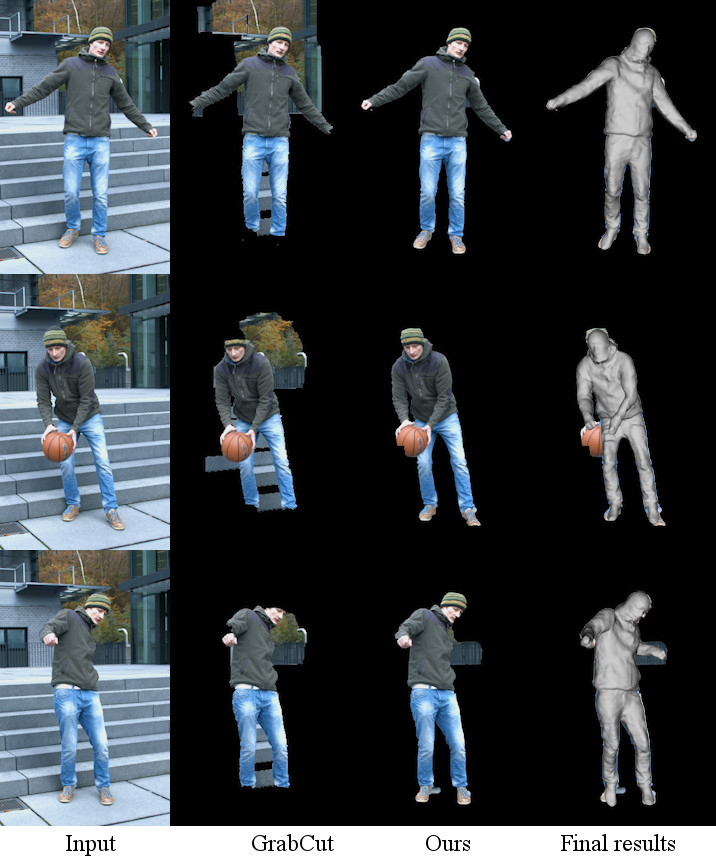}
	\end{center}
	\caption{
		Robustness of our model based silhouette extraction method.
		Our model based silhouette extraction method yields significantly more accurate foreground segmentation results than the default GrabCut method.
		Our correspondence prune strategy improves the robustness of our silhouette based surface refinement to segmentation errors(row 2).
		The failure case, in which the pose estimate is far off, cannot be corrected by our silhouette based refinement (row 3).
	}
	\label{fig:seg_study}
	\vspace{-0.5cm}
\end{figure}

Now we study the effectiveness and robustness of our silhouette extraction method.
As shown in Fig.~\ref{fig:seg_study}, benefiting from the model-based initialization and the motion cue, our model based silhouette extraction method yields significantly more accurate foreground segmentation results than the default GrabCut method.
More segmentation comparison is provided in the supplementary video.
However, segmentation failure occurs inevitably when the actor is occluded by objects (see Fig.~\ref{fig:seg_study} row 2) or the background color is too similar to the foreground.
In these cases, our correspondence prune strategy improves the robustness of our silhouette based surface refinement by checking whether the corresponding points are close enough and whether their normal directions are similar.
If this condition is not met, those correspondences are ignored in the ICP-like iterative refinement, which ensures if segmentation is not good at any part then the mesh is not modified at those parts.
In very rare cases, the estimated pose can be wrong due to occlusion or complicated poses (see Fig.~\ref{fig:seg_study} row 3).
In this case the artifacts cannot be corrected by our silhouette based refinement, since the wrong pose leads to wrong segmentation.
However, our approach instantly recovers once the occluded parts become visible again, see Fig.~\ref{fig:recover} for an example.
This shows that our method does not accumulate errors over time.

\subsection{Comparisons}
\paragraph{Comparison to Monocular Non-rigid Reconstruction}

\begin{figure}[t]
	\begin{center}
		\includegraphics[width=1\linewidth]{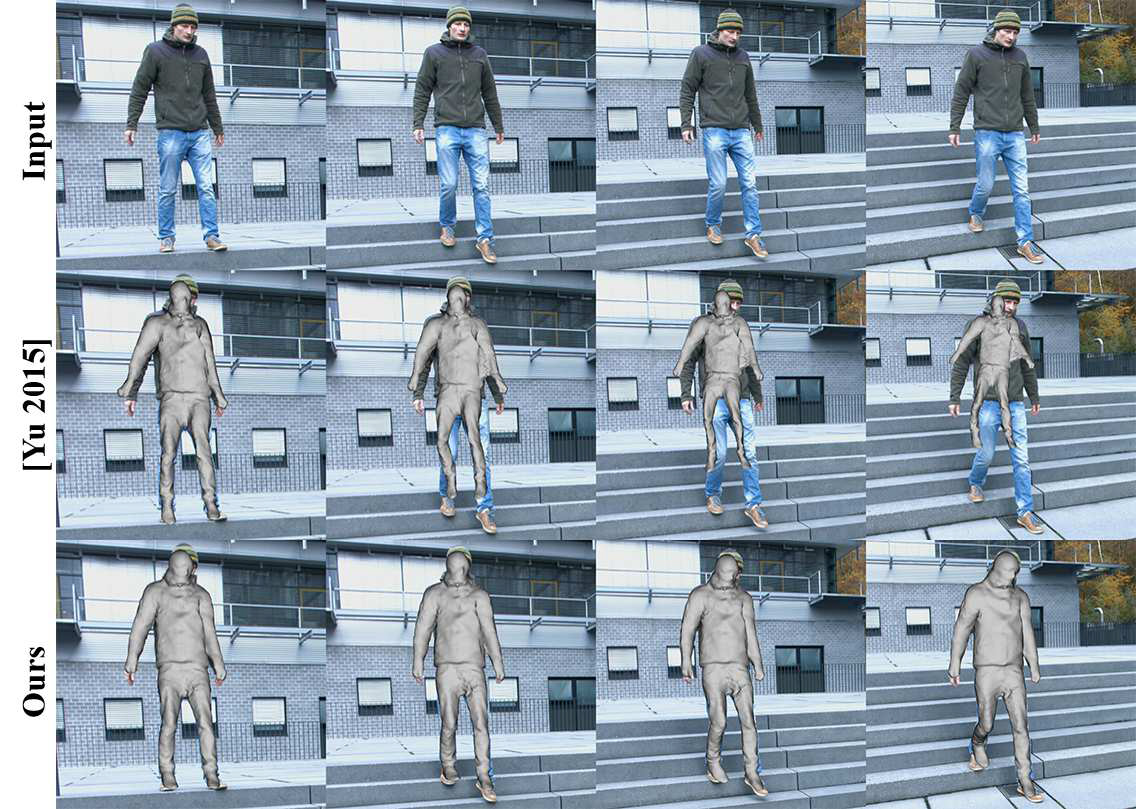}
	\end{center}
	\vspace{-0.4cm}
	\caption{
		Qualitative shape comparison between our approach and the template-based monocular non-rigid tracking approach of \protect\cite{Yu_2015_ICCV}.
		Our approach is able to reconstruct the motion of the complete sequence, while \protect\cite{Yu_2015_ICCV} fails after a few frames.
	}
	\label{fig:comparison_yu}
\end{figure}

\begin{figure}[t]
	\includegraphics[width=\linewidth]{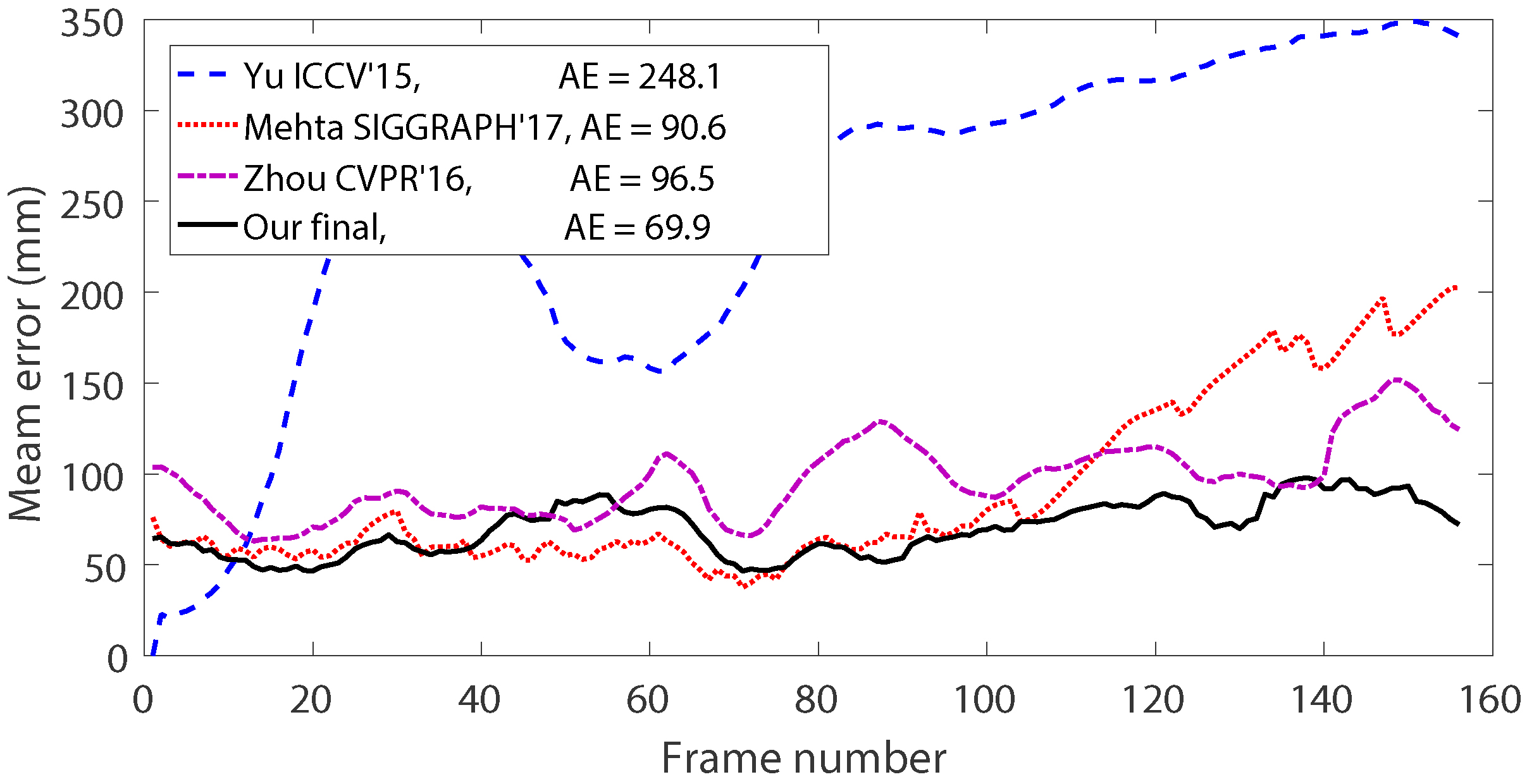}
	\vspace{-0.6cm}
	\caption{
		Quantitative shape comparison on the \emph{Pablo} sequence.
     	Our approach outperforms the template-based tracking approach of \protect\cite{Yu_2015_ICCV} and rigging the template model with respect to the 3D pose estimation results of \protect\cite{VNect_SIGGRAPH2017} and~\protect\cite{zhou2016sparseness}. 
		%
		%
		The average error (AE) in millimeters is given in the legend.
	}
	\label{fig:pablo_yu_quant}
\end{figure}

In Fig.~\ref{fig:comparison_yu}, we provide a qualitative comparison between our approach and the template-based dense monocular non-rigid reconstruction method of~\cite{Yu_2015_ICCV} in terms of the full reconstructed surface.
Their method fails to track the actor motion within a few frames, and is not able to recover afterwards, while our method constantly yields accurate tracking results throughout the entire sequence.
Note that the approach of~\cite{Yu_2015_ICCV} does not rely on a skeleton, and therefore can be applied to general shapes.
However, this comparison confirms the benefits of our shape representation in the specific task of human performance capture.
In addition, we perform a qualitative and quantitative comparison on the \emph{Pablo} sequence.
%
%
As shown in Fig.~\ref{fig:pablo_yu}, the tracking performance of our monocular approach is very close to the multi-view approach of~\cite{robertini2016model} that uses 8 cameras and drastically outperforms the template based monocular approach of~\cite{Yu_2015_ICCV}.
In addition, our approach consistently outperforms theirs in terms of mean vertex error compared to the multi-view reconstructions of~\cite{robertini2016model}, see Fig.~\ref{fig:pablo_yu_quant}.
%
%
%
%

%

\begin{figure}[t]
	\begin{center}
		\includegraphics[width=1\linewidth]{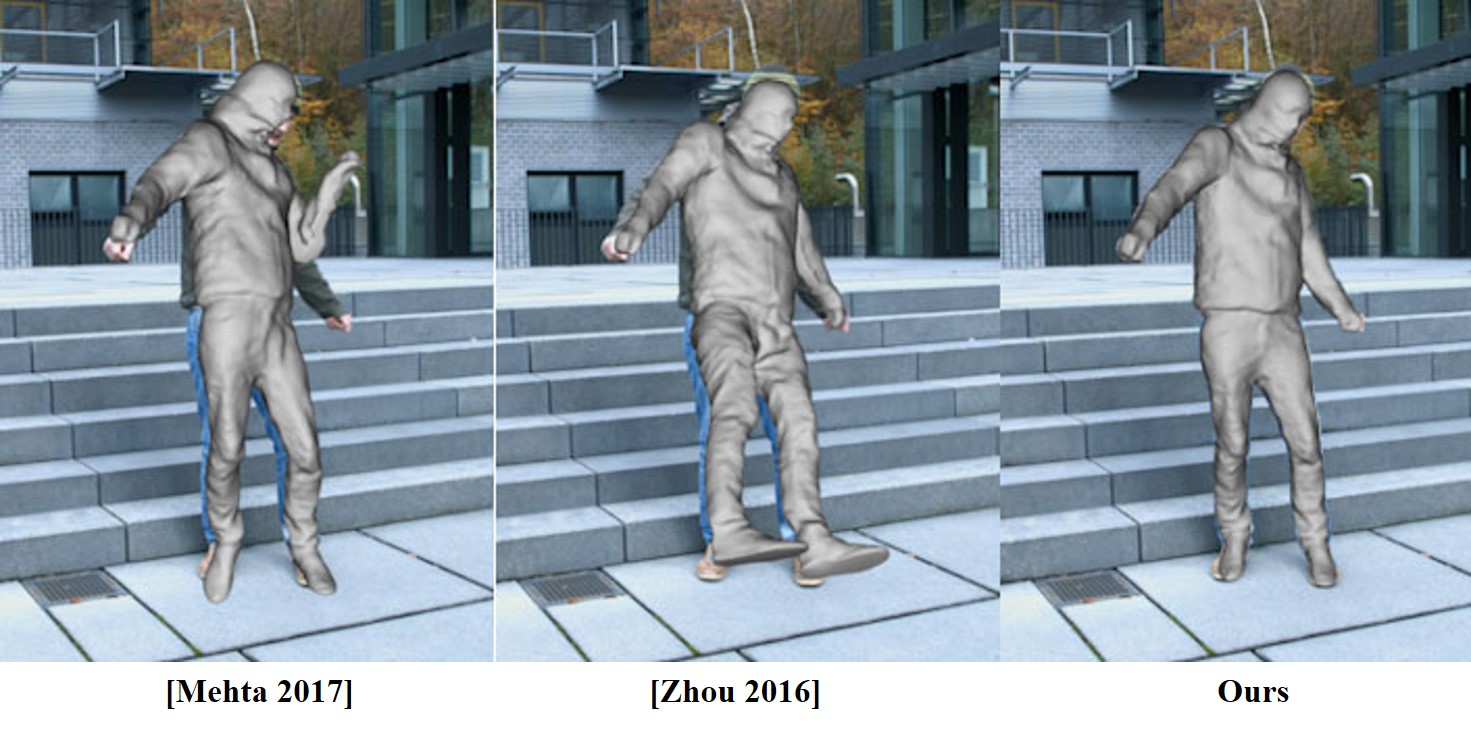}
	\end{center}
	\vspace{-0.4cm}
	\caption{
		Qualitative shape comparison to rigging the template model with respect to the 3D pose estimation results of \protect\cite{VNect_SIGGRAPH2017} and~\protect\cite{zhou2016sparseness}, our results are temporally more stable, of higher quality and better overlay the input. See the accompanying video for more comparisons.
	}
	\label{fig:comparison_helge}
\end{figure}

\paragraph{Comparison to Monocular 3D Joint Estimation}

\begin{figure}[t]
	\begin{center}
		\includegraphics[width=1\linewidth]{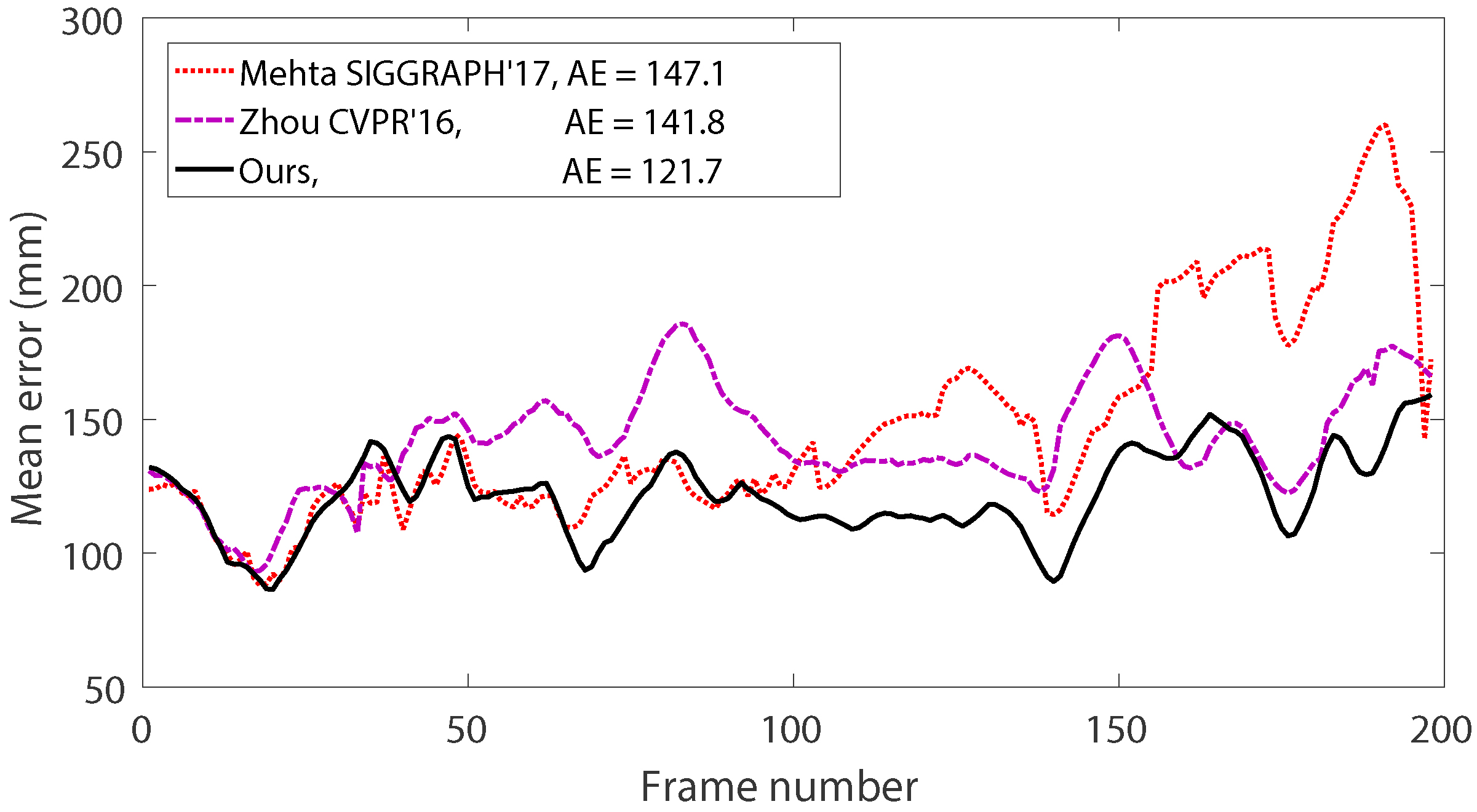}
	\end{center}
	\vspace{-0.4cm}
	\caption{
		Quantitative pose evaluation:
		We compare our average per-joint 3D position error to the approach of \protect\cite{VNect_SIGGRAPH2017} and the 2D-to-3D lifting approach of \protect\cite{zhou2016sparseness}.
		Our approach achieves a consistently lower error leading to higher quality results.
		The average error (AE) in millimeters is given in the legend.
	}
	\vspace{-0.5cm}
	\label{fig:quantitative_joints}
\end{figure}

We also quantitatively compare our batch-based pose estimation method ($E_{\textrm{pose}}$) to the state-of-the-art real-time 3D skeleton tracker of~\cite{VNect_SIGGRAPH2017} and the 2D-to-3D lifting approach of~\cite{zhou2016sparseness}.

We first compare on the basis of joint positions, as the other methods do not reconstruct a deformable surface model.
The ground truth joint locations for this evaluation are provided by the professional multi-view marker-less motion capture software CapturyStudio\footnote{http://www.thecaptury.com/}.
We evaluate the average per-joint 3D error (in millimeters) after similarity transformation for each frame of the \emph{Pablo} sequence. 
As shown in Fig.~\ref{fig:quantitative_joints} (average errors are given in the legend), our batch-based approach that uses 2D and 3D pose detections and fits joint angles of a coherent skeleton model obtains consistently lower errors than theirs.
This lower error in 3D joint positions translates into higher quality and temporally more stable reconstructions.
We also provide a comparison on the publicly available Human3.6M~\cite{h36m_pami} dataset.
To this end, we applied our approch to 4 sequences of Human3.6M, and compare the mean joint error of the 3D skeleton pose results obtained by our batch-optimization to that of~\cite{mehta2016monocular} and~\cite{zhou2016deep}.
To factor out the global pose and bone length scaling, we apply the Procrustes analysis between all pose predictions and the ground truth before computing the errors.
As show in Table~\ref{tab:h3.6m}, our method outperforms these state-of-the-art approaches by a large margin (>3cm improvement).

\begin{table}[t]
	\centering
	\caption{Quantitative pose evaluation on Human3.6M: We compare the mean joint error (in mm) of the 3D skeleton pose results obtained by our batch-optimization to the approach of~\cite{mehta2016monocular} and~\cite{zhou2016deep}.
	Our method outperforms theirs by a large margin.}
	    \begin{tabular}{| r | l | l | l |}
		        \hline
		    & Mehta16 & Zhou16 & \textbf{Ours} \\ \hline
		    S9C2Walking   & 124.93 & 133.37 & \textbf{90.50} \\ \hline
		    S9C2Posing    & 109.36 & 129.28 & \textbf{71.28}  \\ \hline
		    S11C2Greeting & 114.50 & 116.99 & \textbf{75.40} \\ \hline
		    S11C2Sitting  & 112.90 & 110.08 & \textbf{88.70} \\\hline
\end{tabular}
\label{tab:h3.6m}
\end{table}

\begin{figure}[t]
	\begin{center}
		\includegraphics[width=1\linewidth]{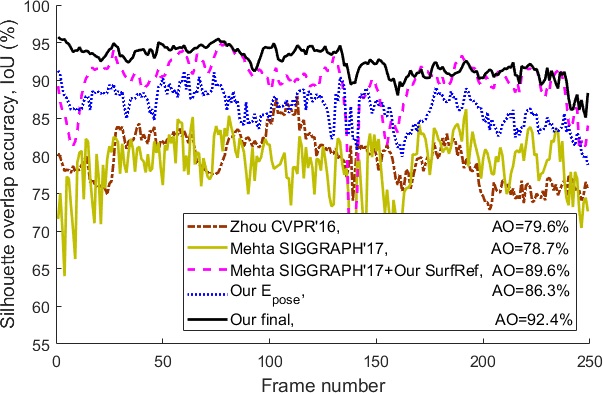}
	\end{center}
	\vspace{-0.1cm}
	\caption{
		Quantitative evaluation:
		We compare our silhouette overlap accuracy to the 2D-to-3D lifting approach of \protect\cite{zhou2016sparseness}, the real-time approach of \protect\cite{VNect_SIGGRAPH2017} and our surface refinement method applied on the results of \protect\cite{VNect_SIGGRAPH2017}.
		Our pose estimation consistently outperforms the existing methods, while our surface refinement further significantly improves the overlap accuracy.
		The average silhouette overlap accuracy (AO) in percentage is given in the legend.
	}
	\vspace{-0.5cm}
	\label{fig:helge_silhouette}
\end{figure}

To further compare our method against the baseline monocular 3D joint estimation methods on surface level, we rigged our template to the pose estimation results of~\cite{VNect_SIGGRAPH2017} and~\cite{zhou2016sparseness}.
Naive rigging exhibits surface artifacts, as shown in Fig.~\ref{fig:comparison_helge} and Fig.~\ref{fig:pablo_yu}, while our approach yields smooth results and improved surface reconstruction quality.
For quantitative comparison, we compute the silhouette overlap accuracy (Intersection over Union, IoU) on the sequence shown in Fig.~\ref{fig:comparison_helge}, based on manually labeled ground truth silhouettes.
As shown in Fig.~\ref{fig:helge_silhouette}, benefiting from our batch-based optimization, our pose estimation consistently outperforms previous state-of-the-art methods, while our surface refinement further significantly improves the overlap accuracy.
To further evaluate our silhouette based surface refinement method, we apply it on the meshes rigged with the pose estimation results of~\cite{VNect_SIGGRAPH2017}.
As shown in Fig.~\ref{fig:helge_silhouette}, the overlap accuracy is improved by 11\% comparing to the skinning results.
Furthermore, Fig.~\ref{fig:pablo_yu_quant} quantitatively shows that our approach outperforms the baseline methods in terms of per-vertex surface reconstruction error.
For more comparisons, we refer to the accompanying video.

\paragraph{Comparison to Stereo Performance Capture}

\begin{figure}
	\begin{center}
		\includegraphics[width=1\linewidth]{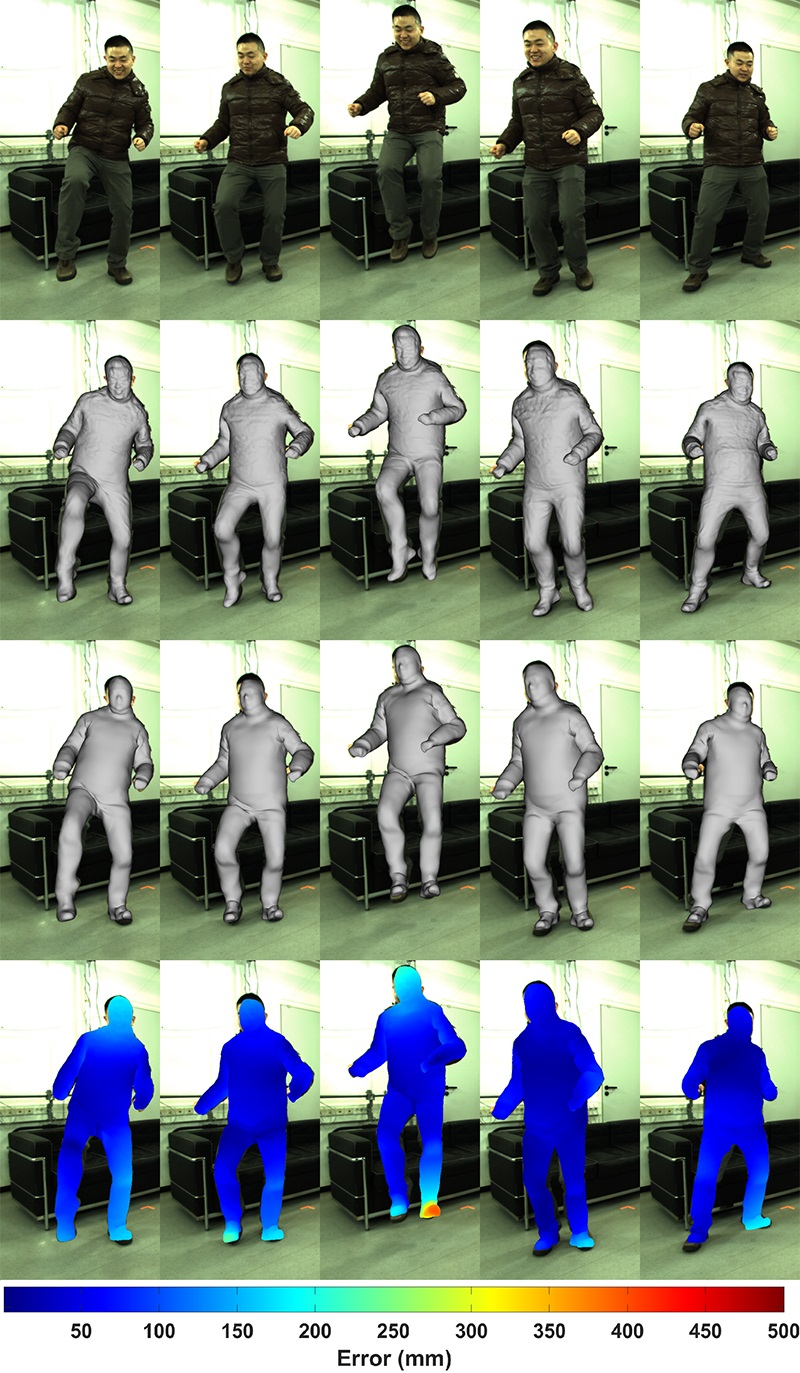}
	\end{center}
	\caption{
		Qualitative comparison to the binocular stereo performance capture approach of \protect\cite{wu2013onset}.
		Our monocular approach obtains comparable quality results without requiring explicit depth cues.
		The per-vertex differences are color coded in the last row.
		Note, we do not employ full BRDF-based shading-based refinement to obtain smoother results.
	}
	\label{fig:comparison_wu}
\end{figure}

We further compare to the stereo-based performance capture approach of~\cite{wu2013onset} on one of their stereo sequences.
This approach leverages explicit depth cues based on binocular stereo and yields high-quality performance capture results.
For comparison, we selected a single camera view (the left camera) to obtain the monocular input video for our approach.
As shown in Fig.~\ref{fig:comparison_wu}, our approach achieves similar accuracy given only monocular input despite the lack of explicit depth cues.
Note, we did not employ full BRDF-based shading-based surface refinement as done in their approach, and thus obtain slightly smoother reconstructions.

\section{Limitations}

\begin{figure}
	\begin{center}
		\includegraphics[width=\linewidth]{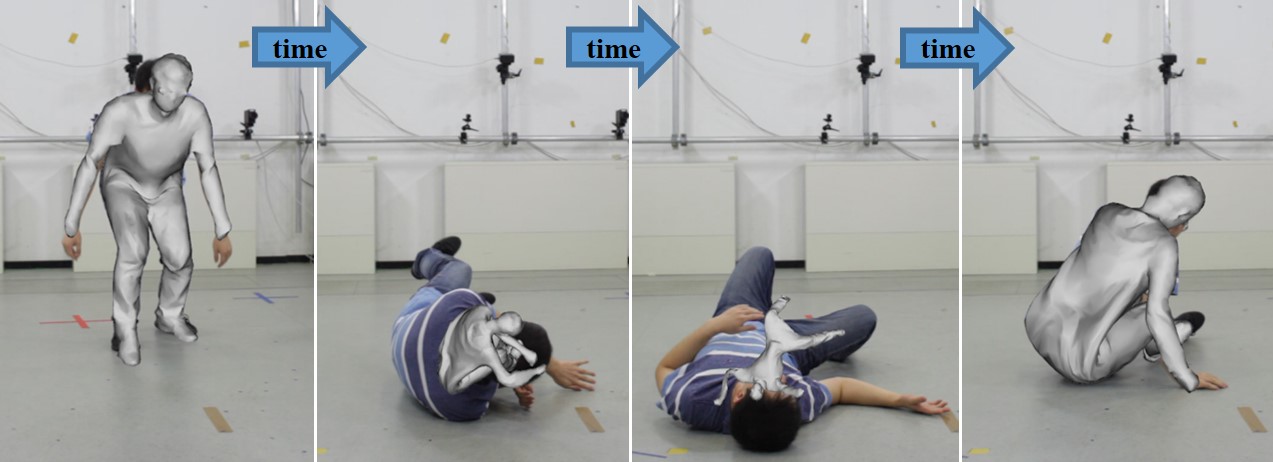}
	\end{center}
	\vspace{-0.3cm}
	\caption{
		Strong occlusion and fast motion can lead to tracking failure, but our approach is able to instantly recover as soon as the occluded parts become visible again.
	}
	\label{fig:recover}
	\vspace{-0.5cm}
\end{figure}

We have demonstrated compelling performance capture results \penalty700\ given just monocular video input.
Nevertheless our approach is subject to the following limitations, which can be addressed in future work:
1) Currently, our approach, similar to previous performance capture approaches, requires a person-specific actor rig built in a pre-processing step. However, note that our template can be automatically generated from a video following a circular path around the static actor, which can be recorded within only half a minute. After this pre-process, our approach is fully automatic. The automatic extraction of such a rig from a monocular video sequence containing general motion is currently an unsolved problem, but first progress given only a sparse set of views \cite{rhodin2016general,bogo2016smpl} has been made.
%
%
%
%
%
%
2) Strong occlusion in combination with fast motion can still lead to tracking failure in our extremely challenging monocular setting.
%
%
Nevertheless, our approach is able to instantly recover due to the discriminative joint detections as soon as the occluded parts become visible again, see Fig.~\ref{fig:recover}.
3) The capturing of feet is less robust in our results. 
Existing datasets for 2D human pose estimation only contain annotation for the major human joints, and unfortunately not for the feet.
Since our model is pre-trained on these 2D datasets it is less robust on feet detection.
Further, since the foot usually has a depth (from toe to the heel) from the camera perspective,  wrong correspondences can be generated while assigning correspondence between the silhouette and the mesh boundary, e.g. a point on the toe being matched to a point on the heel, unless the initial foot position is sufficiently accurate.
However, this issue can be alleviate in future works by incorporating better foot annotations in the training data.
4) Topological changes of complex garments, e.g. opening a jacket, will lead to tracking failure, since only the outer layer of the surface is reconstructed during the template reconstruction.
This can be resolved by reconstructing a multi-layer template.
%
%
%
We would like to remind the reader of the profound difficulty of the monocular reconstruction setting. Despite several remaining limitations, we believe to have taken an important step in monocular performance capture that will inspire follow-up works.



\section{Conclusion}

We have presented the first approach for automatic temporally coherent marker-less human performance capture from a monocular video.
The ambiguities of the underconstrained monocular reconstruction problem are tackled by leveraging sparse 2D and 3D joint detections  and a low dimensional motion prior in a joint optimization problem over a batch of frames.
%
%
The tracked surface geometry is refined based on fully automatically extracted silhouettes to enable medium-scale non-rigid alignment.
We demonstrated compelling monocular reconstruction results that enable exciting applications such as video editing and free viewpoint video previously impossible using single RGB video.

We believe our approach is a significant step to make marker-less monocular performance capture viable.
In the future, a further improved and real-time solution to this challenging problem would have big implications 
%
for a broad range of applications in not only computer animation, visual effects and free-viewpoint video, but also other fields such as medicine or biomechanics.

%

\bibliographystyle{ACM-Reference-Format}
\bibliography{submission} 


\begin{thebibliography}{00}


\ifx \showCODEN    \undefined \def \showCODEN     #1{\unskip}     \fi
\ifx \showDOI      \undefined \def \showDOI       #1{#1}\fi
\ifx \showISBNx    \undefined \def \showISBNx     #1{\unskip}     \fi
\ifx \showISBNxiii \undefined \def \showISBNxiii  #1{\unskip}     \fi
\ifx \showISSN     \undefined \def \showISSN      #1{\unskip}     \fi
\ifx \showLCCN     \undefined \def \showLCCN      #1{\unskip}     \fi
\ifx \shownote     \undefined \def \shownote      #1{#1}          \fi
\ifx \showarticletitle \undefined \def \showarticletitle #1{#1}   \fi
\ifx \showURL      \undefined \def \showURL       {\relax}        \fi
\providecommand\bibfield[2]{#2}
\providecommand\bibinfo[2]{#2}
\providecommand\natexlab[1]{#1}
\providecommand\showeprint[2][]{arXiv:#2}

\bibitem[\protect\citeauthoryear{Akhter and Black}{Akhter and Black}{2015}]%
        {akhter2015poseconditioned}
\bibfield{author}{\bibinfo{person}{Ijaz Akhter} {and}
  \bibinfo{person}{Michael~J Black}.} \bibinfo{year}{2015}\natexlab{}.
\newblock \showarticletitle{Pose-conditioned joint angle limits for 3D human
  pose reconstruction}. In \bibinfo{booktitle}{{\em IEEE Conference on Computer
  Vision and Pattern Recognition (CVPR)}}. \bibinfo{pages}{1446--1455}.
\newblock


\bibitem[\protect\citeauthoryear{Andriluka, Pishchulin, Gehler, and
  Schiele}{Andriluka et~al\mbox{.}}{2014}]%
        {andriluka2014human}
\bibfield{author}{\bibinfo{person}{Mykhaylo Andriluka}, \bibinfo{person}{Leonid
  Pishchulin}, \bibinfo{person}{Peter Gehler}, {and} \bibinfo{person}{Bernt
  Schiele}.} \bibinfo{year}{2014}\natexlab{}.
\newblock \showarticletitle{{2D Human Pose Estimation: New Benchmark and State
  of the Art Analysis}}. In \bibinfo{booktitle}{{\em IEEE Conference on
  Computer Vision and Pattern Recognition (CVPR)}}.
\newblock


\bibitem[\protect\citeauthoryear{Anguelov, Srinivasan, Koller, Thrun, Rodgers,
  and Davis}{Anguelov et~al\mbox{.}}{2005}]%
        {anguelov2005scape}
\bibfield{author}{\bibinfo{person}{Dragomir Anguelov}, \bibinfo{person}{Praveen
  Srinivasan}, \bibinfo{person}{Daphne Koller}, \bibinfo{person}{Sebastian
  Thrun}, \bibinfo{person}{Jim Rodgers}, {and} \bibinfo{person}{James Davis}.}
  \bibinfo{year}{2005}\natexlab{}.
\newblock \showarticletitle{{SCAPE: Shape Completion and Animation of People}}.
\newblock \bibinfo{journal}{{\em ACM Transactions on Graphics\/}}
  \bibinfo{volume}{24}, \bibinfo{number}{3} (\bibinfo{year}{2005}),
  \bibinfo{pages}{408--416}.
\newblock


\bibitem[\protect\citeauthoryear{Balan, Sigal, Black, Davis, and
  Haussecker}{Balan et~al\mbox{.}}{2007}]%
        {balan2007detailed}
\bibfield{author}{\bibinfo{person}{Alexandru~O Balan}, \bibinfo{person}{Leonid
  Sigal}, \bibinfo{person}{Michael~J Black}, \bibinfo{person}{James~E Davis},
  {and} \bibinfo{person}{Horst~W Haussecker}.} \bibinfo{year}{2007}\natexlab{}.
\newblock \showarticletitle{Detailed human shape and pose from images}. In
  \bibinfo{booktitle}{{\em IEEE Conference on Computer Vision and Pattern
  Recognition (CVPR)}}. \bibinfo{pages}{1--8}.
\newblock


\bibitem[\protect\citeauthoryear{Bartoli, Gérard, Chadebecq, Collins, and
  Pizarro}{Bartoli et~al\mbox{.}}{2015}]%
        {Bartoli2015}
\bibfield{author}{\bibinfo{person}{A. Bartoli}, \bibinfo{person}{Y. Gérard},
  \bibinfo{person}{F. Chadebecq}, \bibinfo{person}{T. Collins}, {and}
  \bibinfo{person}{D. Pizarro}.} \bibinfo{year}{2015}\natexlab{}.
\newblock \showarticletitle{Shape-from-Template}.
\newblock \bibinfo{journal}{{\em IEEE Transactions on Pattern Analysis and
  Machine Intelligence\/}} \bibinfo{volume}{37}, \bibinfo{number}{10}
  (\bibinfo{date}{Oct} \bibinfo{year}{2015}), \bibinfo{pages}{2099--2118}.
\newblock
\showISSN{0162-8828}
\showDOI{%
\url{https://doi.org/10.1109/TPAMI.2015.2392759}}


\bibitem[\protect\citeauthoryear{Bogo, Black, Loper, and Romero}{Bogo
  et~al\mbox{.}}{2015}]%
        {bogo2015detailed}
\bibfield{author}{\bibinfo{person}{Federica Bogo}, \bibinfo{person}{Michael~J.
  Black}, \bibinfo{person}{Matthew Loper}, {and} \bibinfo{person}{Javier
  Romero}.} \bibinfo{year}{2015}\natexlab{}.
\newblock \showarticletitle{Detailed Full-Body Reconstructions of Moving People
  from Monocular {RGB-D} Sequences}. In \bibinfo{booktitle}{{\em International
  Conference on Computer Vision (ICCV)}}. \bibinfo{pages}{2300--2308}.
\newblock


\bibitem[\protect\citeauthoryear{Bogo, Kanazawa, Lassner, Gehler, Romero, and
  Black}{Bogo et~al\mbox{.}}{2016}]%
        {bogo2016smpl}
\bibfield{author}{\bibinfo{person}{Federica Bogo}, \bibinfo{person}{Angjoo
  Kanazawa}, \bibinfo{person}{Christoph Lassner}, \bibinfo{person}{Peter
  Gehler}, \bibinfo{person}{Javier Romero}, {and} \bibinfo{person}{Michael~J.
  Black}.} \bibinfo{year}{2016}\natexlab{}.
\newblock \showarticletitle{Keep it {SMPL}: Automatic Estimation of {3D} Human
  Pose and Shape from a Single Image}. In \bibinfo{booktitle}{{\em European
  Conference on Computer Vision (ECCV)}}.
\newblock


\bibitem[\protect\citeauthoryear{Bradley, Popa, Sheffer, Heidrich, and
  Boubekeur}{Bradley et~al\mbox{.}}{2008}]%
        {bradley2008markerless}
\bibfield{author}{\bibinfo{person}{Derek Bradley}, \bibinfo{person}{Tiberiu
  Popa}, \bibinfo{person}{Alla Sheffer}, \bibinfo{person}{Wolfgang Heidrich},
  {and} \bibinfo{person}{Tamy Boubekeur}.} \bibinfo{year}{2008}\natexlab{}.
\newblock \showarticletitle{Markerless garment capture}. In
  \bibinfo{booktitle}{{\em ACM Transactions on Graphics (TOG)}},
  Vol.~\bibinfo{volume}{27}. ACM, \bibinfo{pages}{99}.
\newblock


\bibitem[\protect\citeauthoryear{Bray, Kohli, and Torr}{Bray
  et~al\mbox{.}}{2006}]%
        {bray2006posecut}
\bibfield{author}{\bibinfo{person}{Matthieu Bray}, \bibinfo{person}{Pushmeet
  Kohli}, {and} \bibinfo{person}{Philip~HS Torr}.}
  \bibinfo{year}{2006}\natexlab{}.
\newblock \showarticletitle{Posecut: Simultaneous segmentation and 3d pose
  estimation of humans using dynamic graph-cuts}. In \bibinfo{booktitle}{{\em
  European conference on computer vision}}. Springer,
  \bibinfo{pages}{642--655}.
\newblock


\bibitem[\protect\citeauthoryear{Brox, Rosenhahn, Cremers, and Seidel}{Brox
  et~al\mbox{.}}{2006}]%
        {brox2006high}
\bibfield{author}{\bibinfo{person}{Thomas Brox}, \bibinfo{person}{Bodo
  Rosenhahn}, \bibinfo{person}{Daniel Cremers}, {and}
  \bibinfo{person}{Hans-Peter Seidel}.} \bibinfo{year}{2006}\natexlab{}.
\newblock \showarticletitle{High accuracy optical flow serves 3-D pose
  tracking: exploiting contour and flow based constraints}. In
  \bibinfo{booktitle}{{\em European Conference on Computer Vision}}. Springer,
  \bibinfo{pages}{98--111}.
\newblock


\bibitem[\protect\citeauthoryear{Brox, Rosenhahn, Gall, and Cremers}{Brox
  et~al\mbox{.}}{2010}]%
        {brox2010combined}
\bibfield{author}{\bibinfo{person}{Thomas Brox}, \bibinfo{person}{Bodo
  Rosenhahn}, \bibinfo{person}{Juergen Gall}, {and} \bibinfo{person}{Daniel
  Cremers}.} \bibinfo{year}{2010}\natexlab{}.
\newblock \showarticletitle{Combined region and motion-based 3D tracking of
  rigid and articulated objects}.
\newblock \bibinfo{journal}{{\em IEEE Transactions on Pattern Analysis and
  Machine Intelligence\/}} \bibinfo{volume}{32}, \bibinfo{number}{3}
  (\bibinfo{year}{2010}), \bibinfo{pages}{402--415}.
\newblock


\bibitem[\protect\citeauthoryear{Cagniart, Boyer, and Ilic}{Cagniart
  et~al\mbox{.}}{2010}]%
        {cagniart2010free}
\bibfield{author}{\bibinfo{person}{Cedric Cagniart}, \bibinfo{person}{Edmond
  Boyer}, {and} \bibinfo{person}{Slobodan Ilic}.}
  \bibinfo{year}{2010}\natexlab{}.
\newblock \showarticletitle{Free-form mesh tracking: a patch-based approach}.
  In \bibinfo{booktitle}{{\em Computer Vision and Pattern Recognition (CVPR),
  2010 IEEE Conference on}}. IEEE, \bibinfo{pages}{1339--1346}.
\newblock


\bibitem[\protect\citeauthoryear{Carranza, Theobalt, Magnor, and
  Seidel}{Carranza et~al\mbox{.}}{2003}]%
        {Carranza:2003}
\bibfield{author}{\bibinfo{person}{Joel Carranza}, \bibinfo{person}{Christian
  Theobalt}, \bibinfo{person}{Marcus~A. Magnor}, {and}
  \bibinfo{person}{Hans-Peter Seidel}.} \bibinfo{year}{2003}\natexlab{}.
\newblock \showarticletitle{Free-viewpoint Video of Human Actors}.
\newblock \bibinfo{journal}{{\em ACM Trans. Graph.\/}} \bibinfo{volume}{22},
  \bibinfo{number}{3} (\bibinfo{date}{July} \bibinfo{year}{2003}).
\newblock


\bibitem[\protect\citeauthoryear{Chen, Kim, and Cipolla}{Chen
  et~al\mbox{.}}{2010}]%
        {chen2010inferring}
\bibfield{author}{\bibinfo{person}{Yu Chen}, \bibinfo{person}{Tae-Kyun Kim},
  {and} \bibinfo{person}{Roberto Cipolla}.} \bibinfo{year}{2010}\natexlab{}.
\newblock \showarticletitle{Inferring 3D shapes and deformations from single
  views}. In \bibinfo{booktitle}{{\em European Conference on Computer Vision}}.
  Springer, \bibinfo{pages}{300--313}.
\newblock


\bibitem[\protect\citeauthoryear{Collet, Chuang, Sweeney, Gillett, Evseev,
  Calabrese, Hoppe, Kirk, and Sullivan}{Collet et~al\mbox{.}}{2015}]%
        {collet2015high}
\bibfield{author}{\bibinfo{person}{Alvaro Collet}, \bibinfo{person}{Ming
  Chuang}, \bibinfo{person}{Pat Sweeney}, \bibinfo{person}{Don Gillett},
  \bibinfo{person}{Dennis Evseev}, \bibinfo{person}{David Calabrese},
  \bibinfo{person}{Hugues Hoppe}, \bibinfo{person}{Adam Kirk}, {and}
  \bibinfo{person}{Steve Sullivan}.} \bibinfo{year}{2015}\natexlab{}.
\newblock \showarticletitle{High-quality streamable free-viewpoint video}.
\newblock \bibinfo{journal}{{\em ACM Transactions on Graphics (TOG)\/}}
  \bibinfo{volume}{34}, \bibinfo{number}{4} (\bibinfo{year}{2015}),
  \bibinfo{pages}{69}.
\newblock


\bibitem[\protect\citeauthoryear{De~Aguiar, Stoll, Theobalt, Ahmed, Seidel, and
  Thrun}{De~Aguiar et~al\mbox{.}}{2008}]%
        {de2008performance}
\bibfield{author}{\bibinfo{person}{Edilson De~Aguiar}, \bibinfo{person}{Carsten
  Stoll}, \bibinfo{person}{Christian Theobalt}, \bibinfo{person}{Naveed Ahmed},
  \bibinfo{person}{Hans-Peter Seidel}, {and} \bibinfo{person}{Sebastian
  Thrun}.} \bibinfo{year}{2008}\natexlab{}.
\newblock \showarticletitle{Performance capture from sparse multi-view video}.
  In \bibinfo{booktitle}{{\em ACM Transactions on Graphics (TOG)}},
  Vol.~\bibinfo{volume}{27}. ACM, \bibinfo{pages}{98}.
\newblock


\bibitem[\protect\citeauthoryear{Dou, Fuchs, and Frahm}{Dou
  et~al\mbox{.}}{2013}]%
        {dou2013scanning}
\bibfield{author}{\bibinfo{person}{Mingsong Dou}, \bibinfo{person}{Henry
  Fuchs}, {and} \bibinfo{person}{Jan-Michael Frahm}.}
  \bibinfo{year}{2013}\natexlab{}.
\newblock \showarticletitle{Scanning and tracking dynamic objects with
  commodity depth cameras}. In \bibinfo{booktitle}{{\em Mixed and Augmented
  Reality (ISMAR), 2013 IEEE International Symposium on}}. IEEE,
  \bibinfo{pages}{99--106}.
\newblock


\bibitem[\protect\citeauthoryear{Dou, Khamis, Degtyarev, Davidson, Fanello,
  Kowdle, Escolano, Rhemann, Kim, Taylor, et~al\mbox{.}}{Dou
  et~al\mbox{.}}{2016}]%
        {dou2016fusion4d}
\bibfield{author}{\bibinfo{person}{Mingsong Dou}, \bibinfo{person}{Sameh
  Khamis}, \bibinfo{person}{Yury Degtyarev}, \bibinfo{person}{Philip Davidson},
  \bibinfo{person}{Sean~Ryan Fanello}, \bibinfo{person}{Adarsh Kowdle},
  \bibinfo{person}{Sergio~Orts Escolano}, \bibinfo{person}{Christoph Rhemann},
  \bibinfo{person}{David Kim}, \bibinfo{person}{Jonathan Taylor},
  {et~al\mbox{.}}} \bibinfo{year}{2016}\natexlab{}.
\newblock \showarticletitle{Fusion4d: Real-time performance capture of
  challenging scenes}.
\newblock \bibinfo{journal}{{\em ACM Transactions on Graphics (TOG)\/}}
  \bibinfo{volume}{35}, \bibinfo{number}{4} (\bibinfo{year}{2016}),
  \bibinfo{pages}{114}.
\newblock


\bibitem[\protect\citeauthoryear{Elhayek, de~Aguiar, Jain, Tompson, Pishchulin,
  Andriluka, Bregler, Schiele, and Theobalt}{Elhayek et~al\mbox{.}}{2015}]%
        {elhayek2015efficient}
\bibfield{author}{\bibinfo{person}{Ahmed Elhayek}, \bibinfo{person}{Edilson de
  Aguiar}, \bibinfo{person}{Arjun Jain}, \bibinfo{person}{Jonathan Tompson},
  \bibinfo{person}{Leonid Pishchulin}, \bibinfo{person}{Micha Andriluka},
  \bibinfo{person}{Chris Bregler}, \bibinfo{person}{Bernt Schiele}, {and}
  \bibinfo{person}{Christian Theobalt}.} \bibinfo{year}{2015}\natexlab{}.
\newblock \showarticletitle{Efficient {ConvNet}-based marker-less motion
  capture in general scenes with a low number of cameras}. In
  \bibinfo{booktitle}{{\em IEEE Conference on Computer Vision and Pattern
  Recognition (CVPR)}}. \bibinfo{pages}{3810--3818}.
\newblock


\bibitem[\protect\citeauthoryear{Gall, Stoll, De~Aguiar, Theobalt, Rosenhahn,
  and Seidel}{Gall et~al\mbox{.}}{2009}]%
        {gall2009motion}
\bibfield{author}{\bibinfo{person}{Juergen Gall}, \bibinfo{person}{Carsten
  Stoll}, \bibinfo{person}{Edilson De~Aguiar}, \bibinfo{person}{Christian
  Theobalt}, \bibinfo{person}{Bodo Rosenhahn}, {and}
  \bibinfo{person}{Hans-Peter Seidel}.} \bibinfo{year}{2009}\natexlab{}.
\newblock \showarticletitle{Motion capture using joint skeleton tracking and
  surface estimation}. In \bibinfo{booktitle}{{\em Computer Vision and Pattern
  Recognition, 2009. CVPR 2009. IEEE Conference on}}. IEEE,
  \bibinfo{pages}{1746--1753}.
\newblock


\bibitem[\protect\citeauthoryear{Garg, Roussos, and Agapito}{Garg
  et~al\mbox{.}}{2013}]%
        {Garg_2013_CVPR}
\bibfield{author}{\bibinfo{person}{R. Garg}, \bibinfo{person}{A. Roussos},
  {and} \bibinfo{person}{L. Agapito}.} \bibinfo{year}{2013}\natexlab{}.
\newblock \showarticletitle{Dense Variational Reconstruction of Non-rigid
  Surfaces from Monocular Video}. In \bibinfo{booktitle}{{\em 2013 IEEE
  Conference on Computer Vision and Pattern Recognition}}.
  \bibinfo{pages}{1272--1279}.
\newblock
\showISSN{1063-6919}
\showDOI{%
\url{https://doi.org/10.1109/CVPR.2013.168}}


\bibitem[\protect\citeauthoryear{Garrido, Zollhoefer, Casas, Valgaerts,
  Varanasi, Perez, and Theobalt}{Garrido et~al\mbox{.}}{2016}]%
        {Garrido:2016}
\bibfield{author}{\bibinfo{person}{Pablo Garrido}, \bibinfo{person}{Michael
  Zollhoefer}, \bibinfo{person}{Dan Casas}, \bibinfo{person}{Levi Valgaerts},
  \bibinfo{person}{Kiran Varanasi}, \bibinfo{person}{Patrick Perez}, {and}
  \bibinfo{person}{Christian Theobalt}.} \bibinfo{year}{2016}\natexlab{}.
\newblock \showarticletitle{Reconstruction of Personalized 3D Face Rigs from
  Monocular Video}.
\newblock  \bibinfo{volume}{35}, \bibinfo{number}{3} (\bibinfo{year}{2016}),
  \bibinfo{pages}{28:1--28:15}.
\newblock


\bibitem[\protect\citeauthoryear{Grest, Herzog, and Koch}{Grest
  et~al\mbox{.}}{2005}]%
        {grest2005human}
\bibfield{author}{\bibinfo{person}{Daniel Grest}, \bibinfo{person}{Dennis
  Herzog}, {and} \bibinfo{person}{Reinhard Koch}.}
  \bibinfo{year}{2005}\natexlab{}.
\newblock \showarticletitle{Human model fitting from monocular posture images}.
  In \bibinfo{booktitle}{{\em Proc. of VMV}}.
\newblock


\bibitem[\protect\citeauthoryear{Guan, Weiss, B{\u{a}}lan, and Black}{Guan
  et~al\mbox{.}}{2009}]%
        {guan2009estimating}
\bibfield{author}{\bibinfo{person}{Peng Guan}, \bibinfo{person}{Alexander
  Weiss}, \bibinfo{person}{Alexandru~O B{\u{a}}lan}, {and}
  \bibinfo{person}{Michael~J Black}.} \bibinfo{year}{2009}\natexlab{}.
\newblock \showarticletitle{Estimating human shape and pose from a single
  image}. In \bibinfo{booktitle}{{\em ICCV}}. \bibinfo{pages}{1381--1388}.
\newblock


\bibitem[\protect\citeauthoryear{Guo, Xu, Wang, Liu, and Dai}{Guo
  et~al\mbox{.}}{2015}]%
        {Guo:2015}
\bibfield{author}{\bibinfo{person}{Kaiwen Guo}, \bibinfo{person}{Feng Xu},
  \bibinfo{person}{Yangang Wang}, \bibinfo{person}{Yebin Liu}, {and}
  \bibinfo{person}{Qionghai Dai}.} \bibinfo{year}{2015}\natexlab{}.
\newblock \showarticletitle{Robust Non-rigid Motion Tracking and Surface
  Reconstruction Using L0 Regularization}. In \bibinfo{booktitle}{{\em
  Proceedings of the 2015 IEEE International Conference on Computer Vision
  (ICCV)}} {\em (\bibinfo{series}{ICCV '15})}. \bibinfo{pages}{3083--3091}.
\newblock


\bibitem[\protect\citeauthoryear{Hasler, Ackermann, Rosenhahn, Thorm{\"a}hlen,
  and Seidel}{Hasler et~al\mbox{.}}{2010}]%
        {hasler2010multilinear}
\bibfield{author}{\bibinfo{person}{Nils Hasler}, \bibinfo{person}{Hanno
  Ackermann}, \bibinfo{person}{Bodo Rosenhahn}, \bibinfo{person}{Thorsten
  Thorm{\"a}hlen}, {and} \bibinfo{person}{Hans-Peter Seidel}.}
  \bibinfo{year}{2010}\natexlab{}.
\newblock \showarticletitle{Multilinear pose and body shape estimation of
  dressed subjects from image sets}. In \bibinfo{booktitle}{{\em Computer
  Vision and Pattern Recognition (CVPR), 2010 IEEE Conference on}}. IEEE,
  \bibinfo{pages}{1823--1830}.
\newblock


\bibitem[\protect\citeauthoryear{He, Zhang, Ren, and Sun}{He
  et~al\mbox{.}}{2016}]%
        {he2016deep}
\bibfield{author}{\bibinfo{person}{Kaiming He}, \bibinfo{person}{Xiangyu
  Zhang}, \bibinfo{person}{Shaoqing Ren}, {and} \bibinfo{person}{Jian Sun}.}
  \bibinfo{year}{2016}\natexlab{}.
\newblock \showarticletitle{Deep Residual Learning for Image Recognition}. In
  \bibinfo{booktitle}{{\em IEEE Conference on Computer Vision and Pattern
  Recognition (CVPR)}}.
\newblock


\bibitem[\protect\citeauthoryear{Helten, Muller, Seidel, and Theobalt}{Helten
  et~al\mbox{.}}{2013}]%
        {Helten:2013}
\bibfield{author}{\bibinfo{person}{Thomas Helten}, \bibinfo{person}{Meinard
  Muller}, \bibinfo{person}{Hans-Peter Seidel}, {and}
  \bibinfo{person}{Christian Theobalt}.} \bibinfo{year}{2013}\natexlab{}.
\newblock \showarticletitle{Real-Time Body Tracking with One Depth Camera and
  Inertial Sensors}. In \bibinfo{booktitle}{{\em The IEEE International
  Conference on Computer Vision (ICCV)}}.
\newblock


\bibitem[\protect\citeauthoryear{Huang, Bogo, Lassner, Kanazawa, Gehler,
  Romero, Akhter, and Black}{Huang et~al\mbox{.}}{2017}]%
        {MuVS:3DV:2017}
\bibfield{author}{\bibinfo{person}{Yinghao Huang}, \bibinfo{person}{Federica
  Bogo}, \bibinfo{person}{Christoph Lassner}, \bibinfo{person}{Angjoo
  Kanazawa}, \bibinfo{person}{Peter~V. Gehler}, \bibinfo{person}{Javier
  Romero}, \bibinfo{person}{Ijaz Akhter}, {and} \bibinfo{person}{Michael~J.
  Black}.} \bibinfo{year}{2017}\natexlab{}.
\newblock \showarticletitle{Towards Accurate Marker-less Human Shape and Pose
  Estimation over Time}. In \bibinfo{booktitle}{{\em International Conference
  on 3D Vision (3DV)}}.
\newblock


\bibitem[\protect\citeauthoryear{Innmann, Zollh{\"o}fer, Nie{\ss}ner, Theobalt,
  and Stamminger}{Innmann et~al\mbox{.}}{2016}]%
        {innmann2016volume}
\bibfield{author}{\bibinfo{person}{Matthias Innmann}, \bibinfo{person}{Michael
  Zollh{\"o}fer}, \bibinfo{person}{Matthias Nie{\ss}ner},
  \bibinfo{person}{Christian Theobalt}, {and} \bibinfo{person}{Marc
  Stamminger}.} \bibinfo{year}{2016}\natexlab{}.
\newblock \showarticletitle{{VolumeDeform: Real-time Volumetric Non-rigid
  Reconstruction}}.
\newblock  (\bibinfo{date}{October} \bibinfo{year}{2016}), 17.
\newblock


\bibitem[\protect\citeauthoryear{Ionescu, Carreira, and Sminchisescu}{Ionescu
  et~al\mbox{.}}{2014a}]%
        {ionescu2014iterated}
\bibfield{author}{\bibinfo{person}{Catalin Ionescu}, \bibinfo{person}{Joao
  Carreira}, {and} \bibinfo{person}{Cristian Sminchisescu}.}
  \bibinfo{year}{2014}\natexlab{a}.
\newblock \showarticletitle{Iterated second-order label sensitive pooling for
  3d human pose estimation}. In \bibinfo{booktitle}{{\em IEEE Conference on
  Computer Vision and Pattern Recognition (CVPR)}}.
  \bibinfo{pages}{1661--1668}.
\newblock


\bibitem[\protect\citeauthoryear{Ionescu, Papava, Olaru, and
  Sminchisescu}{Ionescu et~al\mbox{.}}{2014b}]%
        {h36m_pami}
\bibfield{author}{\bibinfo{person}{Catalin Ionescu}, \bibinfo{person}{Dragos
  Papava}, \bibinfo{person}{Vlad Olaru}, {and} \bibinfo{person}{Cristian
  Sminchisescu}.} \bibinfo{year}{2014}\natexlab{b}.
\newblock \showarticletitle{Human3.6M: Large Scale Datasets and Predictive
  Methods for 3D Human Sensing in Natural Environments}.
\newblock \bibinfo{journal}{{\em IEEE Transactions on Pattern Analysis and
  Machine Intelligence\/}} \bibinfo{volume}{36}, \bibinfo{number}{7}
  (\bibinfo{date}{jul} \bibinfo{year}{2014}), \bibinfo{pages}{1325--1339}.
\newblock


\bibitem[\protect\citeauthoryear{Jain, Thorm\"{a}hlen, Seidel, and
  Theobalt}{Jain et~al\mbox{.}}{2010}]%
        {jain2010movie}
\bibfield{author}{\bibinfo{person}{Arjun Jain}, \bibinfo{person}{Thorsten
  Thorm\"{a}hlen}, \bibinfo{person}{Hans-Peter Seidel}, {and}
  \bibinfo{person}{Christian Theobalt}.} \bibinfo{year}{2010}\natexlab{}.
\newblock \showarticletitle{{MovieReshape}: Tracking and Reshaping of Humans in
  Videos}.
\newblock \bibinfo{journal}{{\em ACM Transactions on Graphics\/}}
  \bibinfo{volume}{29}, \bibinfo{number}{5} (\bibinfo{year}{2010}).
\newblock
\showDOI{%
\url{https://doi.org/10.1145/1866158.1866174}}


\bibitem[\protect\citeauthoryear{Jain, Tompson, LeCun, and Bregler}{Jain
  et~al\mbox{.}}{2014}]%
        {jain2014modeep}
\bibfield{author}{\bibinfo{person}{Arjun Jain}, \bibinfo{person}{Jonathan
  Tompson}, \bibinfo{person}{Yann LeCun}, {and} \bibinfo{person}{Christoph
  Bregler}.} \bibinfo{year}{2014}\natexlab{}.
\newblock \showarticletitle{Modeep: A deep learning framework using motion
  features for human pose estimation}. In \bibinfo{booktitle}{{\em Asian
  Conference on Computer Vision (ACCV)}}. Springer, \bibinfo{pages}{302--315}.
\newblock


\bibitem[\protect\citeauthoryear{Johnson and Everingham}{Johnson and
  Everingham}{2011}]%
        {johnson2011learning}
\bibfield{author}{\bibinfo{person}{Sam Johnson} {and} \bibinfo{person}{Mark
  Everingham}.} \bibinfo{year}{2011}\natexlab{}.
\newblock \showarticletitle{Learning Effective Human Pose Estimation from
  Inaccurate Annotation}. In \bibinfo{booktitle}{{\em Proceedings of IEEE
  Conference on Computer Vision and Pattern Recognition}}.
\newblock


\bibitem[\protect\citeauthoryear{Kavan, Collins, \v{Z}\'{a}ra, and
  O'Sullivan}{Kavan et~al\mbox{.}}{2007}]%
        {Kavan:2007}
\bibfield{author}{\bibinfo{person}{Ladislav Kavan}, \bibinfo{person}{Steven
  Collins}, \bibinfo{person}{Ji\v{r}\'{\i} \v{Z}\'{a}ra}, {and}
  \bibinfo{person}{Carol O'Sullivan}.} \bibinfo{year}{2007}\natexlab{}.
\newblock \showarticletitle{Skinning with Dual Quaternions}. In
  \bibinfo{booktitle}{{\em Proceedings of the 2007 Symposium on Interactive 3D
  Graphics and Games}} {\em (\bibinfo{series}{I3D '07})}.
\newblock


\bibitem[\protect\citeauthoryear{Lewis, Cordner, and Fong}{Lewis
  et~al\mbox{.}}{2000}]%
        {Lewis:2000}
\bibfield{author}{\bibinfo{person}{J.~P. Lewis}, \bibinfo{person}{Matt
  Cordner}, {and} \bibinfo{person}{Nickson Fong}.}
  \bibinfo{year}{2000}\natexlab{}.
\newblock \showarticletitle{Pose Space Deformation: A Unified Approach to Shape
  Interpolation and Skeleton-driven Deformation}. In \bibinfo{booktitle}{{\em
  Proceedings of the 27th Annual Conference on Computer Graphics and
  Interactive Techniques}} {\em (\bibinfo{series}{SIGGRAPH '00})}.
  \bibinfo{pages}{165--172}.
\newblock


\bibitem[\protect\citeauthoryear{Li, Adams, Guibas, and Pauly}{Li
  et~al\mbox{.}}{2009}]%
        {Li2009}
\bibfield{author}{\bibinfo{person}{Hao Li}, \bibinfo{person}{Bart Adams},
  \bibinfo{person}{Leonidas~J. Guibas}, {and} \bibinfo{person}{Mark Pauly}.}
  \bibinfo{year}{2009}\natexlab{}.
\newblock \bibinfo{title}{{Robust single-view geometry and motion
  reconstruction}}.
\newblock   (\bibinfo{year}{2009}).
\newblock
\showISBNx{978-1-60558-858-2}
\showISSN{07300301}
\showDOI{%
\url{https://doi.org/10.1145/1618452.1618521}}


\bibitem[\protect\citeauthoryear{Li and Chan}{Li and Chan}{2014}]%
        {li2014threed}
\bibfield{author}{\bibinfo{person}{Sijin Li} {and} \bibinfo{person}{Antoni~B
  Chan}.} \bibinfo{year}{2014}\natexlab{}.
\newblock \showarticletitle{3d human pose estimation from monocular images with
  deep convolutional neural network}. In \bibinfo{booktitle}{{\em Asian
  Conference on Computer Vision (ACCV)}}. \bibinfo{pages}{332--347}.
\newblock


\bibitem[\protect\citeauthoryear{Li, Zhang, and Chan}{Li et~al\mbox{.}}{2015}]%
        {li2015maximummargin}
\bibfield{author}{\bibinfo{person}{Sijin Li}, \bibinfo{person}{Weichen Zhang},
  {and} \bibinfo{person}{Antoni~B Chan}.} \bibinfo{year}{2015}\natexlab{}.
\newblock \showarticletitle{Maximum-margin structured learning with deep
  networks for 3d human pose estimation}. In \bibinfo{booktitle}{{\em IEEE
  International Conference on Computer Vision (ICCV)}}.
  \bibinfo{pages}{2848--2856}.
\newblock


\bibitem[\protect\citeauthoryear{Liu, Stoll, Gall, Seidel, and Theobalt}{Liu
  et~al\mbox{.}}{2011}]%
        {liu2011markerless}
\bibfield{author}{\bibinfo{person}{Yebin Liu}, \bibinfo{person}{Carsten Stoll},
  \bibinfo{person}{Juergen Gall}, \bibinfo{person}{Hans-Peter Seidel}, {and}
  \bibinfo{person}{Christian Theobalt}.} \bibinfo{year}{2011}\natexlab{}.
\newblock \showarticletitle{Markerless motion capture of interacting characters
  using multi-view image segmentation}. In \bibinfo{booktitle}{{\em Computer
  Vision and Pattern Recognition (CVPR), 2011 IEEE Conference on}}. IEEE,
  \bibinfo{pages}{1249--1256}.
\newblock


\bibitem[\protect\citeauthoryear{Loper, Mahmood, and Black}{Loper
  et~al\mbox{.}}{2014}]%
        {loper2014mosh}
\bibfield{author}{\bibinfo{person}{Matthew Loper}, \bibinfo{person}{Naureen
  Mahmood}, {and} \bibinfo{person}{Michael~J Black}.}
  \bibinfo{year}{2014}\natexlab{}.
\newblock \showarticletitle{{MoSh}: Motion and shape capture from sparse
  markers}.
\newblock \bibinfo{journal}{{\em ACM Transactions on Graphics\/}}
  \bibinfo{volume}{33}, \bibinfo{number}{6} (\bibinfo{year}{2014}),
  \bibinfo{pages}{220}.
\newblock


\bibitem[\protect\citeauthoryear{Loper, Mahmood, Romero, Pons-Moll, and
  Black}{Loper et~al\mbox{.}}{2015}]%
        {loper2015smpl}
\bibfield{author}{\bibinfo{person}{Matthew Loper}, \bibinfo{person}{Naureen
  Mahmood}, \bibinfo{person}{Javier Romero}, \bibinfo{person}{Gerard
  Pons-Moll}, {and} \bibinfo{person}{Michael~J Black}.}
  \bibinfo{year}{2015}\natexlab{}.
\newblock \showarticletitle{{SMPL: A skinned multi-person linear model}}.
\newblock \bibinfo{journal}{{\em ACM Transactions on Graphics (TOG)\/}}
  \bibinfo{volume}{34}, \bibinfo{number}{6} (\bibinfo{year}{2015}).
\newblock


\bibitem[\protect\citeauthoryear{Matusik, Buehler, Raskar, Gortler, and
  McMillan}{Matusik et~al\mbox{.}}{2000}]%
        {matusik2000image}
\bibfield{author}{\bibinfo{person}{Wojciech Matusik}, \bibinfo{person}{Chris
  Buehler}, \bibinfo{person}{Ramesh Raskar}, \bibinfo{person}{Steven~J
  Gortler}, {and} \bibinfo{person}{Leonard McMillan}.}
  \bibinfo{year}{2000}\natexlab{}.
\newblock \showarticletitle{Image-based visual hulls}. In
  \bibinfo{booktitle}{{\em Proceedings of the 27th annual conference on
  Computer graphics and interactive techniques}}. ACM Press/Addison-Wesley
  Publishing Co., \bibinfo{pages}{369--374}.
\newblock


\bibitem[\protect\citeauthoryear{Mehta, Rhodin, Casas, Sotnychenko, Xu, and
  Theobalt}{Mehta et~al\mbox{.}}{2016}]%
        {mehta2016monocular}
\bibfield{author}{\bibinfo{person}{Dushyant Mehta}, \bibinfo{person}{Helge
  Rhodin}, \bibinfo{person}{Dan Casas}, \bibinfo{person}{Oleksandr
  Sotnychenko}, \bibinfo{person}{Weipeng Xu}, {and} \bibinfo{person}{Christian
  Theobalt}.} \bibinfo{year}{2016}\natexlab{}.
\newblock \showarticletitle{Monocular 3D Human Pose Estimation Using Transfer
  Learning and Improved CNN Supervision}.
\newblock \bibinfo{journal}{{\em arXiv preprint arXiv:1611.09813\/}}
  (\bibinfo{year}{2016}).
\newblock


\bibitem[\protect\citeauthoryear{Mehta, Sridhar, Sotnychenko, Rhodin, Shafiei,
  Seidel, Xu, Casas, and Theobalt}{Mehta et~al\mbox{.}}{2017}]%
        {VNect_SIGGRAPH2017}
\bibfield{author}{\bibinfo{person}{Dushyant Mehta}, \bibinfo{person}{Srinath
  Sridhar}, \bibinfo{person}{Oleksandr Sotnychenko}, \bibinfo{person}{Helge
  Rhodin}, \bibinfo{person}{Mohammad Shafiei}, \bibinfo{person}{Hans-Peter
  Seidel}, \bibinfo{person}{Weipeng Xu}, \bibinfo{person}{Dan Casas}, {and}
  \bibinfo{person}{Christian Theobalt}.} \bibinfo{year}{2017}\natexlab{}.
\newblock \showarticletitle{VNect: Real-time 3D Human Pose Estimation with a
  Single RGB Camera}.
\newblock \bibinfo{journal}{{\em ACM Transactions on Graphics (Proceedings of
  SIGGRAPH 2017)\/}} \bibinfo{volume}{36}, \bibinfo{number}{4}, 14.
\newblock


\bibitem[\protect\citeauthoryear{Mori and Malik}{Mori and Malik}{2006}]%
        {mori2006recovering}
\bibfield{author}{\bibinfo{person}{Greg Mori} {and} \bibinfo{person}{Jitendra
  Malik}.} \bibinfo{year}{2006}\natexlab{}.
\newblock \showarticletitle{Recovering 3d human body configurations using shape
  contexts}.
\newblock \bibinfo{journal}{{\em IEEE Transactions on Pattern Analysis and
  Machine Intelligence (TPAMI)\/}} \bibinfo{volume}{28}, \bibinfo{number}{7}
  (\bibinfo{year}{2006}), \bibinfo{pages}{1052--1062}.
\newblock


\bibitem[\protect\citeauthoryear{Mustafa, Kim, Guillemaut, and Hilton}{Mustafa
  et~al\mbox{.}}{2015}]%
        {mustafa2015iccv}
\bibfield{author}{\bibinfo{person}{Armin Mustafa}, \bibinfo{person}{Hansung
  Kim}, \bibinfo{person}{Jean-Yves Guillemaut}, {and} \bibinfo{person}{Adrian
  Hilton}.} \bibinfo{year}{2015}\natexlab{}.
\newblock \showarticletitle{General Dynamic Scene Reconstruction from Multiple
  View Video}. In \bibinfo{booktitle}{{\em ICCV}}.
\newblock


\bibitem[\protect\citeauthoryear{Newcombe, Fox, and Seitz}{Newcombe
  et~al\mbox{.}}{2015}]%
        {Newcombe_2015_CVPR}
\bibfield{author}{\bibinfo{person}{Richard~A. Newcombe},
  \bibinfo{person}{Dieter Fox}, {and} \bibinfo{person}{Steven~M. Seitz}.}
  \bibinfo{year}{2015}\natexlab{}.
\newblock \showarticletitle{DynamicFusion: Reconstruction and Tracking of
  Non-Rigid Scenes in Real-Time}. In \bibinfo{booktitle}{{\em The IEEE
  Conference on Computer Vision and Pattern Recognition (CVPR)}}.
\newblock


\bibitem[\protect\citeauthoryear{Newell, Yang, and Deng}{Newell
  et~al\mbox{.}}{2016}]%
        {newell2016stacked}
\bibfield{author}{\bibinfo{person}{Alejandro Newell}, \bibinfo{person}{Kaiyu
  Yang}, {and} \bibinfo{person}{Jia Deng}.} \bibinfo{year}{2016}\natexlab{}.
\newblock \showarticletitle{Stacked hourglass networks for human pose
  estimation}.
\newblock \bibinfo{journal}{{\em arXiv preprint arXiv:1603.06937\/}}
  (\bibinfo{year}{2016}).
\newblock


\bibitem[\protect\citeauthoryear{Park, Shiratori, Matthews, and Sheikh}{Park
  et~al\mbox{.}}{2015}]%
        {Hyun2015}
\bibfield{author}{\bibinfo{person}{Hyun~Soo Park}, \bibinfo{person}{Takaaki
  Shiratori}, \bibinfo{person}{Iain Matthews}, {and} \bibinfo{person}{Yaser
  Sheikh}.} \bibinfo{year}{2015}\natexlab{}.
\newblock \showarticletitle{{3D Trajectory Reconstruction under Perspective
  Projection}}.
\newblock \bibinfo{journal}{{\em International Journal of Computer Vision\/}}
  \bibinfo{volume}{115}, \bibinfo{number}{2} (\bibinfo{year}{2015}),
  \bibinfo{pages}{115--135}.
\newblock
\showISBNx{0920-5691}
\showISSN{15731405}
\showDOI{%
\url{https://doi.org/10.1007/s11263-015-0804-2}}


\bibitem[\protect\citeauthoryear{Pavlakos, Zhou, Derpanis, and
  Daniilidis}{Pavlakos et~al\mbox{.}}{2016}]%
        {pavlakos2016coarse}
\bibfield{author}{\bibinfo{person}{Georgios Pavlakos}, \bibinfo{person}{Xiaowei
  Zhou}, \bibinfo{person}{Konstantinos~G Derpanis}, {and}
  \bibinfo{person}{Kostas Daniilidis}.} \bibinfo{year}{2016}\natexlab{}.
\newblock \showarticletitle{Coarse-to-Fine Volumetric Prediction for
  Single-Image 3D Human Pose}.
\newblock \bibinfo{journal}{{\em arXiv preprint arXiv:1611.07828\/}}
  (\bibinfo{year}{2016}).
\newblock


\bibitem[\protect\citeauthoryear{Pishchulin, Insafutdinov, Tang, Andres,
  Andriluka, Gehler, and Schiele}{Pishchulin et~al\mbox{.}}{2016}]%
        {pishchulin2016deepcut}
\bibfield{author}{\bibinfo{person}{Leonid Pishchulin}, \bibinfo{person}{Eldar
  Insafutdinov}, \bibinfo{person}{Siyu Tang}, \bibinfo{person}{Bjoern Andres},
  \bibinfo{person}{Mykhaylo Andriluka}, \bibinfo{person}{Peter Gehler}, {and}
  \bibinfo{person}{Bernt Schiele}.} \bibinfo{year}{2016}\natexlab{}.
\newblock \showarticletitle{DeepCut: Joint Subset Partition and Labeling for
  Multi Person Pose Estimation}. In \bibinfo{booktitle}{{\em IEEE Conference on
  Computer Vision and Pattern Recognition (CVPR)}}.
\newblock


\bibitem[\protect\citeauthoryear{Pl{\"a}nkers and Fua}{Pl{\"a}nkers and
  Fua}{2001}]%
        {plankers2001tracking}
\bibfield{author}{\bibinfo{person}{Ralf Pl{\"a}nkers} {and}
  \bibinfo{person}{Pascal Fua}.} \bibinfo{year}{2001}\natexlab{}.
\newblock \showarticletitle{Tracking and modeling people in video sequences}.
\newblock \bibinfo{journal}{{\em Computer Vision and Image Understanding\/}}
  \bibinfo{volume}{81}, \bibinfo{number}{3} (\bibinfo{year}{2001}),
  \bibinfo{pages}{285--302}.
\newblock


\bibitem[\protect\citeauthoryear{Rhodin, Robertini, Casas, Richardt, Seidel,
  and Theobalt}{Rhodin et~al\mbox{.}}{2016}]%
        {rhodin2016general}
\bibfield{author}{\bibinfo{person}{Helge Rhodin}, \bibinfo{person}{Nadia
  Robertini}, \bibinfo{person}{Dan Casas}, \bibinfo{person}{Christian
  Richardt}, \bibinfo{person}{Hans-Peter Seidel}, {and}
  \bibinfo{person}{Christian Theobalt}.} \bibinfo{year}{2016}\natexlab{}.
\newblock \showarticletitle{General Automatic Human Shape and Motion Capture
  Using Volumetric Contour Cues}. In \bibinfo{booktitle}{{\em ECCV}},
  \bibfield{editor}{\bibinfo{person}{Bastian Leibe}, \bibinfo{person}{Jiri
  Matas}, \bibinfo{person}{Nicu Sebe}, {and} \bibinfo{person}{Max Welling}}
  (Eds.). \bibinfo{publisher}{Springer International Publishing},
  \bibinfo{address}{Cham}, \bibinfo{pages}{509--526}.
\newblock


\bibitem[\protect\citeauthoryear{Robertini, Casas, Rhodin, Seidel, and
  Theobalt}{Robertini et~al\mbox{.}}{2016}]%
        {robertini2016model}
\bibfield{author}{\bibinfo{person}{Nadia Robertini}, \bibinfo{person}{Dan
  Casas}, \bibinfo{person}{Helge Rhodin}, \bibinfo{person}{Hans-Peter Seidel},
  {and} \bibinfo{person}{Christian Theobalt}.} \bibinfo{year}{2016}\natexlab{}.
\newblock \showarticletitle{{Model-based Outdoor Performance Capture}}. In
  \bibinfo{booktitle}{{\em International Conference on Computer Vision (3DV)}}.
\newblock


\bibitem[\protect\citeauthoryear{Rogge, Klose, Stengel, Eisemann, and
  Magnor}{Rogge et~al\mbox{.}}{2014}]%
        {rogge2014garment}
\bibfield{author}{\bibinfo{person}{Lorenz Rogge}, \bibinfo{person}{Felix
  Klose}, \bibinfo{person}{Michael Stengel}, \bibinfo{person}{Martin Eisemann},
  {and} \bibinfo{person}{Marcus Magnor}.} \bibinfo{year}{2014}\natexlab{}.
\newblock \showarticletitle{Garment replacement in monocular video sequences}.
\newblock \bibinfo{journal}{{\em ACM Transactions on Graphics (TOG)\/}}
  \bibinfo{volume}{34}, \bibinfo{number}{1} (\bibinfo{year}{2014}),
  \bibinfo{pages}{6}.
\newblock


\bibitem[\protect\citeauthoryear{Rosales and Sclaroff}{Rosales and
  Sclaroff}{2006}]%
        {rosales2006combining}
\bibfield{author}{\bibinfo{person}{R{\'o}mer Rosales} {and}
  \bibinfo{person}{Stan Sclaroff}.} \bibinfo{year}{2006}\natexlab{}.
\newblock \showarticletitle{Combining generative and discriminative models in a
  framework for articulated pose estimation}.
\newblock \bibinfo{journal}{{\em International Journal of Computer Vision\/}}
  \bibinfo{volume}{67}, \bibinfo{number}{3} (\bibinfo{year}{2006}),
  \bibinfo{pages}{251--276}.
\newblock


\bibitem[\protect\citeauthoryear{Rother, Kolmogorov, and Blake}{Rother
  et~al\mbox{.}}{2004}]%
        {rother2004grabcut}
\bibfield{author}{\bibinfo{person}{Carsten Rother}, \bibinfo{person}{Vladimir
  Kolmogorov}, {and} \bibinfo{person}{Andrew Blake}.}
  \bibinfo{year}{2004}\natexlab{}.
\newblock \showarticletitle{Grab{C}ut: Interactive foreground extraction using
  iterated graph cuts}. In \bibinfo{booktitle}{{\em ACM transactions on
  graphics (TOG)}}, Vol.~\bibinfo{volume}{23}. ACM, \bibinfo{pages}{309--314}.
\newblock


\bibitem[\protect\citeauthoryear{Russell, Yu, and Agapito}{Russell
  et~al\mbox{.}}{2014}]%
        {Russell2014}
\bibfield{author}{\bibinfo{person}{Chris Russell}, \bibinfo{person}{Rui Yu},
  {and} \bibinfo{person}{Lourdes Agapito}.} \bibinfo{year}{2014}\natexlab{}.
\newblock \bibinfo{booktitle}{{\em Video Pop-up: Monocular 3D Reconstruction of
  Dynamic Scenes}}.
\newblock \bibinfo{publisher}{Springer International Publishing},
  \bibinfo{address}{Cham}, \bibinfo{pages}{583--598}.
\newblock
\showISBNx{978-3-319-10584-0}
\showDOI{%
\url{https://doi.org/10.1007/978-3-319-10584-0_38}}


\bibitem[\protect\citeauthoryear{Salzmann and Fua}{Salzmann and Fua}{2011}]%
        {Salzmann2011}
\bibfield{author}{\bibinfo{person}{Mathieu Salzmann} {and}
  \bibinfo{person}{Pascal Fua}.} \bibinfo{year}{2011}\natexlab{}.
\newblock \showarticletitle{{Linear local models for monocular reconstruction
  of deformable surfaces}}.
\newblock \bibinfo{journal}{{\em IEEE Transactions on Pattern Analysis and
  Machine Intelligence\/}} \bibinfo{volume}{33}, \bibinfo{number}{5}
  (\bibinfo{year}{2011}), \bibinfo{pages}{931--944}.
\newblock
\showISBNx{978-1-4244-2242-5}
\showISSN{01628828}
\showDOI{%
\url{https://doi.org/10.1109/TPAMI.2010.158}}


\bibitem[\protect\citeauthoryear{Shotton, Fitzgibbon, Cook, Sharp, Finocchio,
  Moore, Kipman, and Blake}{Shotton et~al\mbox{.}}{2011}]%
        {shotton2011real}
\bibfield{author}{\bibinfo{person}{J. Shotton}, \bibinfo{person}{A.
  Fitzgibbon}, \bibinfo{person}{M. Cook}, \bibinfo{person}{T. Sharp},
  \bibinfo{person}{M. Finocchio}, \bibinfo{person}{R. Moore},
  \bibinfo{person}{A. Kipman}, {and} \bibinfo{person}{A. Blake}.}
  \bibinfo{year}{2011}\natexlab{}.
\newblock \showarticletitle{Real-time Human Pose Recognition in Parts from
  Single Depth Images}. In \bibinfo{booktitle}{{\em Proceedings of the 2011
  IEEE Conference on Computer Vision and Pattern Recognition}} {\em
  (\bibinfo{series}{CVPR '11})}. \bibinfo{pages}{1297--1304}.
\newblock


\bibitem[\protect\citeauthoryear{Sidenbladh, Black, and Fleet}{Sidenbladh
  et~al\mbox{.}}{2000}]%
        {sidenbladh2000stochastic}
\bibfield{author}{\bibinfo{person}{Hedvig Sidenbladh},
  \bibinfo{person}{Michael~J. Black}, {and} \bibinfo{person}{David~J. Fleet}.}
  \bibinfo{year}{2000}\natexlab{}.
\newblock \showarticletitle{Stochastic Tracking of 3D Human Figures Using 2D
  Image Motion}. In \bibinfo{booktitle}{{\em ECCV}}. \bibinfo{pages}{702--718}.
\newblock


\bibitem[\protect\citeauthoryear{Sigal, Balan, and Black}{Sigal
  et~al\mbox{.}}{2007}]%
        {sigal2007combined}
\bibfield{author}{\bibinfo{person}{Leonid Sigal}, \bibinfo{person}{Alexandru
  Balan}, {and} \bibinfo{person}{Michael~J Black}.}
  \bibinfo{year}{2007}\natexlab{}.
\newblock \showarticletitle{Combined discriminative and generative articulated
  pose and non-rigid shape estimation}. In \bibinfo{booktitle}{{\em Advances in
  neural information processing systems}}. \bibinfo{pages}{1337--1344}.
\newblock


\bibitem[\protect\citeauthoryear{Simo-Serra, Ramisa, Aleny{\`a}, Torras, and
  Moreno-Noguer}{Simo-Serra et~al\mbox{.}}{2012}]%
        {simo2012single}
\bibfield{author}{\bibinfo{person}{Edgar Simo-Serra}, \bibinfo{person}{Arnau
  Ramisa}, \bibinfo{person}{Guillem Aleny{\`a}}, \bibinfo{person}{Carme
  Torras}, {and} \bibinfo{person}{Francesc Moreno-Noguer}.}
  \bibinfo{year}{2012}\natexlab{}.
\newblock \showarticletitle{Single image 3d human pose estimation from noisy
  observations}. In \bibinfo{booktitle}{{\em IEEE Conference on Computer Vision
  and Pattern Recognition (CVPR)}}. IEEE, \bibinfo{pages}{2673--2680}.
\newblock


\bibitem[\protect\citeauthoryear{Sminchisescu, Kanaujia, and
  Metaxas}{Sminchisescu et~al\mbox{.}}{2006}]%
        {sminchisescu2006learning}
\bibfield{author}{\bibinfo{person}{Cristian Sminchisescu},
  \bibinfo{person}{Atul Kanaujia}, {and} \bibinfo{person}{Dimitris Metaxas}.}
  \bibinfo{year}{2006}\natexlab{}.
\newblock \showarticletitle{Learning joint top-down and bottom-up processes for
  3D visual inference}. In \bibinfo{booktitle}{{\em 2006 IEEE Computer Society
  Conference on Computer Vision and Pattern Recognition (CVPR'06)}},
  Vol.~\bibinfo{volume}{2}. IEEE, \bibinfo{pages}{1743--1752}.
\newblock


\bibitem[\protect\citeauthoryear{Sminchisescu and Triggs}{Sminchisescu and
  Triggs}{2003a}]%
        {sminchisescu2003estimating}
\bibfield{author}{\bibinfo{person}{Cristian Sminchisescu} {and}
  \bibinfo{person}{Bill Triggs}.} \bibinfo{year}{2003}\natexlab{a}.
\newblock \showarticletitle{Estimating articulated human motion with covariance
  scaled sampling}.
\newblock \bibinfo{journal}{{\em The International Journal of Robotics
  Research\/}} \bibinfo{volume}{22}, \bibinfo{number}{6}
  (\bibinfo{year}{2003}), \bibinfo{pages}{371--391}.
\newblock


\bibitem[\protect\citeauthoryear{Sminchisescu and Triggs}{Sminchisescu and
  Triggs}{2003b}]%
        {sminchisescu2003kinematic}
\bibfield{author}{\bibinfo{person}{Cristian Sminchisescu} {and}
  \bibinfo{person}{Bill Triggs}.} \bibinfo{year}{2003}\natexlab{b}.
\newblock \showarticletitle{Kinematic jump processes for monocular 3D human
  tracking}. In \bibinfo{booktitle}{{\em Computer Vision and Pattern
  Recognition, 2003. Proceedings. 2003 IEEE Computer Society Conference on}},
  Vol.~\bibinfo{volume}{1}. IEEE, \bibinfo{pages}{I--69}.
\newblock


\bibitem[\protect\citeauthoryear{Song, Tong, Chang, Yang, Tang, and Zhang}{Song
  et~al\mbox{.}}{2016}]%
        {song20163d}
\bibfield{author}{\bibinfo{person}{Dan Song}, \bibinfo{person}{Ruofeng Tong},
  \bibinfo{person}{Jian Chang}, \bibinfo{person}{Xiaosong Yang},
  \bibinfo{person}{Min Tang}, {and} \bibinfo{person}{Jian~Jun Zhang}.}
  \bibinfo{year}{2016}\natexlab{}.
\newblock \showarticletitle{3D Body Shapes Estimation from Dressed-Human
  Silhouettes}. In \bibinfo{booktitle}{{\em Computer Graphics Forum}},
  Vol.~\bibinfo{volume}{35}. Wiley Online Library, \bibinfo{pages}{147--156}.
\newblock


\bibitem[\protect\citeauthoryear{Sorkine and Alexa}{Sorkine and Alexa}{2007}]%
        {Sorkine:2007}
\bibfield{author}{\bibinfo{person}{Olga Sorkine} {and} \bibinfo{person}{Marc
  Alexa}.} \bibinfo{year}{2007}\natexlab{}.
\newblock \showarticletitle{As-rigid-as-possible Surface Modeling}. In
  \bibinfo{booktitle}{{\em Proceedings of the Fifth Eurographics Symposium on
  Geometry Processing}} {\em (\bibinfo{series}{SGP '07})}.
  \bibinfo{publisher}{Eurographics Association}.
\newblock


\bibitem[\protect\citeauthoryear{Starck and Hilton}{Starck and Hilton}{2007}]%
        {starck2007surface}
\bibfield{author}{\bibinfo{person}{Jonathan Starck} {and}
  \bibinfo{person}{Adrian Hilton}.} \bibinfo{year}{2007}\natexlab{}.
\newblock \showarticletitle{Surface capture for performance-based animation}.
\newblock \bibinfo{journal}{{\em IEEE Computer Graphics and Applications\/}}
  \bibinfo{volume}{27}, \bibinfo{number}{3} (\bibinfo{year}{2007}),
  \bibinfo{pages}{21--31}.
\newblock


\bibitem[\protect\citeauthoryear{Stoll, Hasler, Gall, Seidel, and
  Theobalt}{Stoll et~al\mbox{.}}{2011}]%
        {StollHGST11}
\bibfield{author}{\bibinfo{person}{Carsten Stoll}, \bibinfo{person}{Nils
  Hasler}, \bibinfo{person}{Juergen Gall}, \bibinfo{person}{Hans{-}Peter
  Seidel}, {and} \bibinfo{person}{Christian Theobalt}.}
  \bibinfo{year}{2011}\natexlab{}.
\newblock \showarticletitle{Fast articulated motion tracking using a sums of
  Gaussians body model}. In \bibinfo{booktitle}{{\em {IEEE} International
  Conference on Computer Vision, {ICCV} 2011, Barcelona, Spain, November 6-13,
  2011}}. \bibinfo{pages}{951--958}.
\newblock


\bibitem[\protect\citeauthoryear{Sumner, Schmid, and Pauly}{Sumner
  et~al\mbox{.}}{2007}]%
        {Sumner:2007}
\bibfield{author}{\bibinfo{person}{Robert~W. Sumner}, \bibinfo{person}{Johannes
  Schmid}, {and} \bibinfo{person}{Mark Pauly}.}
  \bibinfo{year}{2007}\natexlab{}.
\newblock \showarticletitle{Embedded Deformation for Shape Manipulation}.
\newblock \bibinfo{journal}{{\em ACM Trans. Graph.\/}} \bibinfo{volume}{26},
  \bibinfo{number}{3} (\bibinfo{date}{July} \bibinfo{year}{2007}).
\newblock


\bibitem[\protect\citeauthoryear{Taylor}{Taylor}{2000}]%
        {taylor2000reconstruction}
\bibfield{author}{\bibinfo{person}{Camillo~J Taylor}.}
  \bibinfo{year}{2000}\natexlab{}.
\newblock \showarticletitle{Reconstruction of articulated objects from point
  correspondences in a single uncalibrated image}. In \bibinfo{booktitle}{{\em
  IEEE Conference on Computer Vision and Pattern Recognition (CVPR)}},
  Vol.~\bibinfo{volume}{1}. \bibinfo{pages}{677--684}.
\newblock


\bibitem[\protect\citeauthoryear{Tekin, Katircioglu, Salzmann, Lepetit, and
  Fua}{Tekin et~al\mbox{.}}{2016}]%
        {tekin2016structured}
\bibfield{author}{\bibinfo{person}{Bugra Tekin}, \bibinfo{person}{Isinsu
  Katircioglu}, \bibinfo{person}{Mathieu Salzmann}, \bibinfo{person}{Vincent
  Lepetit}, {and} \bibinfo{person}{Pascal Fua}.}
  \bibinfo{year}{2016}\natexlab{}.
\newblock \showarticletitle{{Structured Prediction of 3D Human Pose with Deep
  Neural Networks}}. In \bibinfo{booktitle}{{\em British Machine Vision
  Conference (BMVC)}}.
\newblock


\bibitem[\protect\citeauthoryear{Thies, Zollh{\"o}fer, Stamminger, Theobalt,
  and Nie{\ss}ner}{Thies et~al\mbox{.}}{2016}]%
        {thies2016face}
\bibfield{author}{\bibinfo{person}{J. Thies}, \bibinfo{person}{M.
  Zollh{\"o}fer}, \bibinfo{person}{M. Stamminger}, \bibinfo{person}{C.
  Theobalt}, {and} \bibinfo{person}{M. Nie{\ss}ner}.}
  \bibinfo{year}{2016}\natexlab{}.
\newblock \showarticletitle{{Face2Face: Real-time Face Capture and Reenactment
  of RGB Videos}}. In \bibinfo{booktitle}{{\em Proc. Computer Vision and
  Pattern Recognition (CVPR), IEEE}}.
\newblock


\bibitem[\protect\citeauthoryear{Toshev and Szegedy}{Toshev and
  Szegedy}{2014}]%
        {toshev2014deeppose}
\bibfield{author}{\bibinfo{person}{Alexander Toshev} {and}
  \bibinfo{person}{Christian Szegedy}.} \bibinfo{year}{2014}\natexlab{}.
\newblock \showarticletitle{Deeppose: Human pose estimation via deep neural
  networks}. In \bibinfo{booktitle}{{\em Conference on Computer Vision and
  Pattern Recognition (CVPR)}}. \bibinfo{pages}{1653--1660}.
\newblock


\bibitem[\protect\citeauthoryear{Urtasun, Fleet, and Fua}{Urtasun
  et~al\mbox{.}}{2005}]%
        {urtasun2005monocular}
\bibfield{author}{\bibinfo{person}{Raquel Urtasun}, \bibinfo{person}{David~J
  Fleet}, {and} \bibinfo{person}{Pascal Fua}.} \bibinfo{year}{2005}\natexlab{}.
\newblock \showarticletitle{Monocular 3D tracking of the golf swing}. In
  \bibinfo{booktitle}{{\em CVPR}}. \bibinfo{pages}{932--938}.
\newblock


\bibitem[\protect\citeauthoryear{Urtasun, Fleet, and Fua}{Urtasun
  et~al\mbox{.}}{2006}]%
        {urtasun2006temporal}
\bibfield{author}{\bibinfo{person}{Raquel Urtasun}, \bibinfo{person}{David~J
  Fleet}, {and} \bibinfo{person}{Pascal Fua}.} \bibinfo{year}{2006}\natexlab{}.
\newblock \showarticletitle{Temporal motion models for monocular and multiview
  {3D} human body tracking}.
\newblock \bibinfo{journal}{{\em CVIU\/}} \bibinfo{volume}{104},
  \bibinfo{number}{2} (\bibinfo{year}{2006}), \bibinfo{pages}{157--177}.
\newblock


\bibitem[\protect\citeauthoryear{Vlasic, Baran, Matusik, and
  Popovi{\'c}}{Vlasic et~al\mbox{.}}{2008}]%
        {vlasic2008articulated}
\bibfield{author}{\bibinfo{person}{Daniel Vlasic}, \bibinfo{person}{Ilya
  Baran}, \bibinfo{person}{Wojciech Matusik}, {and} \bibinfo{person}{Jovan
  Popovi{\'c}}.} \bibinfo{year}{2008}\natexlab{}.
\newblock \showarticletitle{Articulated mesh animation from multi-view
  silhouettes}. In \bibinfo{booktitle}{{\em ACM Transactions on Graphics
  (TOG)}}, Vol.~\bibinfo{volume}{27}. ACM, \bibinfo{pages}{97}.
\newblock


\bibitem[\protect\citeauthoryear{Vlasic, Peers, Baran, Debevec, Popovi{\'c},
  Rusinkiewicz, and Matusik}{Vlasic et~al\mbox{.}}{2009}]%
        {vlasic2009dynamic}
\bibfield{author}{\bibinfo{person}{Daniel Vlasic}, \bibinfo{person}{Pieter
  Peers}, \bibinfo{person}{Ilya Baran}, \bibinfo{person}{Paul Debevec},
  \bibinfo{person}{Jovan Popovi{\'c}}, \bibinfo{person}{Szymon Rusinkiewicz},
  {and} \bibinfo{person}{Wojciech Matusik}.} \bibinfo{year}{2009}\natexlab{}.
\newblock \showarticletitle{Dynamic shape capture using multi-view photometric
  stereo}.
\newblock \bibinfo{journal}{{\em ACM Transactions on Graphics (TOG)\/}}
  \bibinfo{volume}{28}, \bibinfo{number}{5} (\bibinfo{year}{2009}),
  \bibinfo{pages}{174}.
\newblock


\bibitem[\protect\citeauthoryear{Wang, Wang, Lin, Yuille, and Gao}{Wang
  et~al\mbox{.}}{2014}]%
        {wang2014robust}
\bibfield{author}{\bibinfo{person}{Chunyu Wang}, \bibinfo{person}{Yizhou Wang},
  \bibinfo{person}{Zhouchen Lin}, \bibinfo{person}{Alan~L Yuille}, {and}
  \bibinfo{person}{Wen Gao}.} \bibinfo{year}{2014}\natexlab{}.
\newblock \showarticletitle{Robust estimation of 3d human poses from a single
  image}. In \bibinfo{booktitle}{{\em IEEE Conference on Computer Vision and
  Pattern Recognition (CVPR)}}. \bibinfo{pages}{2361--2368}.
\newblock


\bibitem[\protect\citeauthoryear{Wang, Wei, Vouga, Huang, Ceylan, Medioni, and
  Li}{Wang et~al\mbox{.}}{2016}]%
        {wang2016capturing}
\bibfield{author}{\bibinfo{person}{Ruizhe Wang}, \bibinfo{person}{Lingyu Wei},
  \bibinfo{person}{Etienne Vouga}, \bibinfo{person}{Qixing Huang},
  \bibinfo{person}{Duygu Ceylan}, \bibinfo{person}{Gerard Medioni}, {and}
  \bibinfo{person}{Hao Li}.} \bibinfo{year}{2016}\natexlab{}.
\newblock \showarticletitle{Capturing Dynamic Textured Surfaces of Moving
  Targets}. In \bibinfo{booktitle}{{\em Proceedings of the European Conference
  on Computer Vision ({ECCV})}}.
\newblock


\bibitem[\protect\citeauthoryear{Waschb{\"u}sch, W{\"u}rmlin, Cotting, Sadlo,
  and Gross}{Waschb{\"u}sch et~al\mbox{.}}{2005}]%
        {waschbusch2005scalable}
\bibfield{author}{\bibinfo{person}{Michael Waschb{\"u}sch},
  \bibinfo{person}{Stephan W{\"u}rmlin}, \bibinfo{person}{Daniel Cotting},
  \bibinfo{person}{Filip Sadlo}, {and} \bibinfo{person}{Markus Gross}.}
  \bibinfo{year}{2005}\natexlab{}.
\newblock \showarticletitle{Scalable 3D video of dynamic scenes}.
\newblock \bibinfo{journal}{{\em The Visual Computer\/}} \bibinfo{volume}{21},
  \bibinfo{number}{8-10} (\bibinfo{year}{2005}), \bibinfo{pages}{629--638}.
\newblock


\bibitem[\protect\citeauthoryear{Wei, Ramakrishna, Kanade, and Sheikh}{Wei
  et~al\mbox{.}}{2016}]%
        {wei2016convolutional}
\bibfield{author}{\bibinfo{person}{Shih-En Wei}, \bibinfo{person}{Varun
  Ramakrishna}, \bibinfo{person}{Takeo Kanade}, {and} \bibinfo{person}{Yaser
  Sheikh}.} \bibinfo{year}{2016}\natexlab{}.
\newblock \showarticletitle{{Convolutional Pose Machines}}. In
  \bibinfo{booktitle}{{\em Conference on Computer Vision and Pattern
  Recognition (CVPR)}}.
\newblock


\bibitem[\protect\citeauthoryear{Wei and Chai}{Wei and Chai}{2010}]%
        {wei2010videomocap}
\bibfield{author}{\bibinfo{person}{Xiaolin Wei} {and} \bibinfo{person}{Jinxiang
  Chai}.} \bibinfo{year}{2010}\natexlab{}.
\newblock \showarticletitle{Videomocap: modeling physically realistic human
  motion from monocular video sequences}. In \bibinfo{booktitle}{{\em ACM
  Transactions on Graphics (TOG)}}, Vol.~\bibinfo{volume}{29}. ACM,
  \bibinfo{pages}{42}.
\newblock


\bibitem[\protect\citeauthoryear{Wren, Azarbayejani, Darrell, and
  Pentland}{Wren et~al\mbox{.}}{1997}]%
        {wren1997pfinder}
\bibfield{author}{\bibinfo{person}{Christopher~Richard Wren},
  \bibinfo{person}{Ali Azarbayejani}, \bibinfo{person}{Trevor Darrell}, {and}
  \bibinfo{person}{Alex~Paul Pentland}.} \bibinfo{year}{1997}\natexlab{}.
\newblock \showarticletitle{Pfinder: real-time tracking of the human body}.
\newblock \bibinfo{journal}{{\em IEEE Transactions on Pattern Analysis and
  Machine Intelligence (PAMI)\/}} \bibinfo{volume}{19}, \bibinfo{number}{7}
  (\bibinfo{year}{1997}), \bibinfo{pages}{780--785}.
\newblock


\bibitem[\protect\citeauthoryear{Wu, Stoll, Valgaerts, and Theobalt}{Wu
  et~al\mbox{.}}{2013}]%
        {wu2013onset}
\bibfield{author}{\bibinfo{person}{Chenglei Wu}, \bibinfo{person}{Carsten
  Stoll}, \bibinfo{person}{Levi Valgaerts}, {and} \bibinfo{person}{Christian
  Theobalt}.} \bibinfo{year}{2013}\natexlab{}.
\newblock \showarticletitle{{On-set Performance Capture of Multiple Actors With
  A Stereo Camera}}. In \bibinfo{booktitle}{{\em ACM Transactions on Graphics
  (Proceedings of SIGGRAPH Asia 2013)}}, Vol.~\bibinfo{volume}{32}.
  \bibinfo{pages}{161:1--161:11}.
\newblock
\showDOI{%
\url{https://doi.org/10.1145/2508363.2508418}}


\bibitem[\protect\citeauthoryear{Wu, Varanasi, and Theobalt}{Wu
  et~al\mbox{.}}{2012}]%
        {wu2012full}
\bibfield{author}{\bibinfo{person}{Chenglei Wu}, \bibinfo{person}{Kiran
  Varanasi}, {and} \bibinfo{person}{Christian Theobalt}.}
  \bibinfo{year}{2012}\natexlab{}.
\newblock \showarticletitle{Full body performance capture under uncontrolled
  and varying illumination: A shading-based approach}. In
  \bibinfo{booktitle}{{\em ECCV}}. \bibinfo{pages}{757--770}.
\newblock


\bibitem[\protect\citeauthoryear{Xu, Salzmann, Wang, and Liu}{Xu
  et~al\mbox{.}}{2015}]%
        {xu2015deformable}
\bibfield{author}{\bibinfo{person}{Weipeng Xu}, \bibinfo{person}{Mathieu
  Salzmann}, \bibinfo{person}{Yongtian Wang}, {and} \bibinfo{person}{Yue Liu}.}
  \bibinfo{year}{2015}\natexlab{}.
\newblock \showarticletitle{Deformable 3D Fusion: From Partial Dynamic 3D
  Observations to Complete 4D Models}.
\newblock \bibinfo{journal}{{\em 2015 IEEE International Conference on Computer
  Vision (ICCV)\/}} (\bibinfo{year}{2015}), \bibinfo{pages}{2183--2191}.
\newblock
\showISSN{2380-7504}


\bibitem[\protect\citeauthoryear{Yasin, Iqbal, Kr{\"u}ger, Weber, and
  Gall}{Yasin et~al\mbox{.}}{2016}]%
        {yasin2016dualsource}
\bibfield{author}{\bibinfo{person}{Hashim Yasin}, \bibinfo{person}{Umar Iqbal},
  \bibinfo{person}{Bj{\"o}rn Kr{\"u}ger}, \bibinfo{person}{Andreas Weber},
  {and} \bibinfo{person}{Juergen Gall}.} \bibinfo{year}{2016}\natexlab{}.
\newblock \showarticletitle{{A Dual-Source Approach for 3D Pose Estimation from
  a Single Image}}. In \bibinfo{booktitle}{{\em Conference on Computer Vision
  and Pattern Recognition (CVPR)}}.
\newblock


\bibitem[\protect\citeauthoryear{Ye, Liu, Hasler, Ji, Dai, and Theobalt}{Ye
  et~al\mbox{.}}{2012}]%
        {Ye2012}
\bibfield{author}{\bibinfo{person}{Genzhi Ye}, \bibinfo{person}{Yebin Liu},
  \bibinfo{person}{Nils Hasler}, \bibinfo{person}{Xiangyang Ji},
  \bibinfo{person}{Qionghai Dai}, {and} \bibinfo{person}{Christian Theobalt}.}
  \bibinfo{year}{2012}\natexlab{}.
\newblock \showarticletitle{{Performance capture of interacting characters with
  handheld kinects}}. In \bibinfo{booktitle}{{\em ECCV}},
  Vol.~\bibinfo{volume}{7573 LNCS}. \bibinfo{pages}{828--841}.
\newblock
\showISBNx{9783642337086}
\showISSN{03029743}
\showDOI{%
\url{https://doi.org/10.1007/978-3-642-33709-3_59}}


\bibitem[\protect\citeauthoryear{Yu, Russell, Campbell, and Agapito}{Yu
  et~al\mbox{.}}{2015}]%
        {Yu_2015_ICCV}
\bibfield{author}{\bibinfo{person}{Rui Yu}, \bibinfo{person}{Chris Russell},
  \bibinfo{person}{Neill D.~F. Campbell}, {and} \bibinfo{person}{Lourdes
  Agapito}.} \bibinfo{year}{2015}\natexlab{}.
\newblock \showarticletitle{Direct, Dense, and Deformable: Template-Based
  Non-Rigid 3D Reconstruction From RGB Video}. In \bibinfo{booktitle}{{\em The
  IEEE International Conference on Computer Vision (ICCV)}}.
\newblock


\bibitem[\protect\citeauthoryear{Zhang, Fu, Ye, and Yang}{Zhang
  et~al\mbox{.}}{2014}]%
        {Zhang2014}
\bibfield{author}{\bibinfo{person}{Qing Zhang}, \bibinfo{person}{Bo Fu},
  \bibinfo{person}{Mao Ye}, {and} \bibinfo{person}{Ruigang Yang}.}
  \bibinfo{year}{2014}\natexlab{}.
\newblock \showarticletitle{{Quality Dynamic Human Body Modeling Using a Single
  Low-cost Depth Camera}}. In \bibinfo{booktitle}{{\em CVPR}}.
  \bibinfo{publisher}{IEEE}, \bibinfo{pages}{676--683}.
\newblock


\bibitem[\protect\citeauthoryear{Zhou, Fu, Liu, Cohen-Or, and Han}{Zhou
  et~al\mbox{.}}{2010}]%
        {zhou2010parametric}
\bibfield{author}{\bibinfo{person}{Shizhe Zhou}, \bibinfo{person}{Hongbo Fu},
  \bibinfo{person}{Ligang Liu}, \bibinfo{person}{Daniel Cohen-Or}, {and}
  \bibinfo{person}{Xiaoguang Han}.} \bibinfo{year}{2010}\natexlab{}.
\newblock \showarticletitle{Parametric reshaping of human bodies in images}.
\newblock \bibinfo{journal}{{\em ACM Transactions on Graphics (TOG)\/}}
  \bibinfo{volume}{29}, \bibinfo{number}{4} (\bibinfo{year}{2010}),
  \bibinfo{pages}{126}.
\newblock


\bibitem[\protect\citeauthoryear{Zhou, Leonardos, Hu, and Daniilidis}{Zhou
  et~al\mbox{.}}{2015}]%
        {zhou2015convex}
\bibfield{author}{\bibinfo{person}{Xiaowei Zhou}, \bibinfo{person}{Spyridon
  Leonardos}, \bibinfo{person}{Xiaoyan Hu}, {and} \bibinfo{person}{Kostas
  Daniilidis}.} \bibinfo{year}{2015}\natexlab{}.
\newblock \showarticletitle{{3D shape estimation from 2D landmarks: A convex
  relaxation approach}}. In \bibinfo{booktitle}{{\em IEEE Conference on
  Computer Vision and Pattern Recognition (CVPR)}}.
  \bibinfo{pages}{4447--4455}.
\newblock


\bibitem[\protect\citeauthoryear{Zhou, Sun, Zhang, Liang, and Wei}{Zhou
  et~al\mbox{.}}{2016a}]%
        {zhou2016deep}
\bibfield{author}{\bibinfo{person}{Xingyi Zhou}, \bibinfo{person}{Xiao Sun},
  \bibinfo{person}{Wei Zhang}, \bibinfo{person}{Shuang Liang}, {and}
  \bibinfo{person}{Yichen Wei}.} \bibinfo{year}{2016}\natexlab{a}.
\newblock \showarticletitle{Deep Kinematic Pose Regression}.
\newblock \bibinfo{journal}{{\em arXiv preprint arXiv:1609.05317\/}}
  (\bibinfo{year}{2016}).
\newblock


\bibitem[\protect\citeauthoryear{Zhou, Zhu, Leonardos, Derpanis, and
  Daniilidis}{Zhou et~al\mbox{.}}{2016b}]%
        {zhou2016sparseness}
\bibfield{author}{\bibinfo{person}{Xiaowei Zhou}, \bibinfo{person}{Menglong
  Zhu}, \bibinfo{person}{Spyridon Leonardos}, \bibinfo{person}{Konstantinos~G
  Derpanis}, {and} \bibinfo{person}{Kostas Daniilidis}.}
  \bibinfo{year}{2016}\natexlab{b}.
\newblock \showarticletitle{Sparseness meets deepness: 3D human pose estimation
  from monocular video}. In \bibinfo{booktitle}{{\em Proceedings of the IEEE
  Conference on Computer Vision and Pattern Recognition}}.
  \bibinfo{pages}{4966--4975}.
\newblock


\bibitem[\protect\citeauthoryear{Zollh{\"o}fer, Nie{\ss}ner, Izadi, Rhemann,
  Zach, Fisher, Wu, Fitzgibbon, Loop, Theobalt, and Stamminger}{Zollh{\"o}fer
  et~al\mbox{.}}{2014}]%
        {zollhoefer2014deformable}
\bibfield{author}{\bibinfo{person}{Michael Zollh{\"o}fer},
  \bibinfo{person}{Matthias Nie{\ss}ner}, \bibinfo{person}{Shahram Izadi},
  \bibinfo{person}{Christoph Rhemann}, \bibinfo{person}{Christopher Zach},
  \bibinfo{person}{Matthew Fisher}, \bibinfo{person}{Chenglei Wu},
  \bibinfo{person}{Andrew Fitzgibbon}, \bibinfo{person}{Charles Loop},
  \bibinfo{person}{Christian Theobalt}, {and} \bibinfo{person}{Marc
  Stamminger}.} \bibinfo{year}{2014}\natexlab{}.
\newblock \showarticletitle{Real-time Non-rigid Reconstruction using an RGB-D
  Camera}.
\newblock \bibinfo{journal}{{\em ACM Transactions on Graphics (TOG)\/}}
  \bibinfo{volume}{33}, \bibinfo{number}{4} (\bibinfo{year}{2014}).
\newblock


\end{thebibliography}

\end{document}